%% file: main-ijcv-revised.tex
\let\LaTeXcline\cline
\documentclass[default,iicol]{sn-jnl}
\let\cline\LaTeXcline% Default with double column layout
\input{math_commands.tex}

\usepackage{graphicx}%
\usepackage{multirow}%
\usepackage{amsmath,amssymb,amsfonts}%
\usepackage{amsthm}%
\usepackage{mathrsfs}%
\usepackage[title]{appendix}%
\usepackage{xcolor}%
\usepackage{textcomp}%
\usepackage{manyfoot}%
\usepackage{booktabs}%
\usepackage{algorithm}%
\usepackage{algorithmicx}%
\usepackage{algpseudocode}%
\usepackage{listings}%
\usepackage{caption}
\usepackage{subcaption}

\newtheorem{theorem}{Theorem}%  meant for continuous numbers
\newtheorem{definition}{Definition}%

\newtheorem{thm}{Theorem} 

\newtheorem{lemma}[thm]{Lemma} 
 
\newtheorem{corollary}[thm]{Corollary}

\DeclareMathOperator{\SNR}{SNR}
\newcommand{\ie}{\emph{i.e.}}
\newcommand{\eg}{\emph{e.g.}}
\begin{document}

\title[Learning Rate Curriculum]{Learning Rate Curriculum}


\author[1]{\fnm{Florinel-Alin} \sur{Croitoru}}

\author[1,2]{\fnm{Nicolae-C\u{a}t\u{a}lin} \sur{Ristea}}

\author*[1]{\fnm{Radu Tudor} \sur{Ionescu}}\email{raducu.ionescu@gmail.com}

\author[3]{\fnm{Nicu} \sur{Sebe}}

\affil[1]{\orgdiv{Department of Computer Science},
\orgname{University of Bucharest}, \orgaddress{\street{14 Academiei},
\city{Bucharest},
\postcode{010014},
\country{Romania}}}

\affil[2]{\orgdiv{Faculty of Electronics, Telecommunications, and Information Technology},
\orgname{National University of Science and Technology Politehnica Bucharest}, \orgaddress{\street{313 Splaiul Independentei},
\city{Bucharest}, \postcode{060042}, \country{Romania}}}

\affil[3]{\orgdiv{Department of Information Engineering and Computer Science},
\orgname{University of Trento}, \orgaddress{\street{9 via Sommarive},
\city{Povo-Trento}, \postcode{38123},
\country{Italy}}}

\abstract{
%\vspace{-0.3cm}
Most curriculum learning methods require an approach to sort the data samples by difficulty, which is often cumbersome to perform. In this work, we propose a novel curriculum learning approach termed \textbf{Le}arning \textbf{Ra}te \textbf{C}urriculum (LeRaC), which leverages the use of a different learning rate for each layer of a neural network to create a data-agnostic curriculum during the initial training epochs. More specifically, LeRaC assigns higher learning rates to neural layers closer to the input, gradually decreasing the learning rates as the layers are placed farther away from the input. The learning rates increase at various paces during the first training iterations, until they all reach the same value. From this point on, the neural model is trained as usual. This creates a model-level curriculum learning strategy that does not require sorting the examples by difficulty and is compatible with any neural network, generating higher performance levels regardless of the architecture. We conduct comprehensive experiments on 12 data sets from the computer vision (CIFAR-10, CIFAR-100, Tiny ImageNet, ImageNet-200, Food-101, UTKFace, PASCAL VOC), language (BoolQ, QNLI, RTE) and audio (ESC-50, CREMA-D) domains, considering various convolutional (ResNet-18, Wide-ResNet-50, DenseNet-121, YOLOv5), recurrent (LSTM) and transformer (CvT, BERT, SepTr) architectures. We compare our approach with the conventional training regime, as well as with Curriculum by Smoothing (CBS), a state-of-the-art data-agnostic curriculum learning approach. Unlike CBS, our performance improvements over the standard training regime are consistent across all data sets and models. Furthermore, we significantly surpass CBS in terms of training time (there is no additional cost over the standard training regime for LeRaC). Our code is freely available at: \url{https://github.com/CroitoruAlin/LeRaC}.
%\vspace{-0.6cm}
}
\maketitle

% \setlength{\abovedisplayskip}{3.0pt}
% \setlength{\belowdisplayskip}{3.0pt}

\section{Introduction}

% Machine learning researchers relentlessly strive to improve the performance of AI models. Much of this effort has been directed to the development of novel neural architectures \cite{Brown-NeurIPS-2020,Devlin-NAACL-2019,Dosovitskiy-ICLR-2020,He-CVPR-2016,Khan-ACS-2021,Liu-CVPR-2022,Raffel-JMLR-2020,Wu-ICCV-2021,Zhu-ICLR-2020}, which have grown in size and complexity \cite{Brown-NeurIPS-2020,Raffel-JMLR-2020,Reed-ArXiv-2022} to leverage the availability of increasingly larger data sets. However, we believe the dominant trend to develop deeper and deeper neural networks is not sustainable on the long term. To this end, we turn our attention to an alternative approach to increase performance of deep neural models without growing the size of the respective models. More specifically, we focus on curriculum learning, an approach initially proposed by Bengio \etal~\cite{Bengio-ICML-2009} to train better neural networks by mimicking how humans learn, from easy to hard. 

Curriculum learning \citep{Bengio-ICML-2009} refers to efficiently training effective neural networks by mimicking how humans learn, from easy to hard. As originally introduced by \citet{Bengio-ICML-2009}, curriculum learning is a training procedure that first organizes the examples in their increasing order of difficulty, then starts the training of the neural network on the easiest examples, gradually adding increasingly more difficult examples along the way, until all training examples are fed into the network. The success of the approach relies in avoiding imposing the learning of very difficult examples right from the beginning, instead guiding the model on the right path through the imposed curriculum. This type of curriculum is later referred to as data-level curriculum learning \citep{Soviany-IJCV-2022}. Indeed, \citet{Soviany-IJCV-2022} identified several types of curriculum learning approaches in the literature, dividing them into four categories based on the components involved in the definition of machine learning given by \citet{Mitchell-MH-1997}. The four categories are: data-level curriculum (examples are presented from easy to hard), model-level curriculum (the modeling capacity of the network is gradually increased), task-level curriculum (the complexity of the learning task is increased during training), objective-level curriculum (the model optimizes towards an increasingly more complex objective). While data-level curriculum is the most natural and direct way to employ curriculum learning, its main disadvantage is that it requires a way to determine the difficulty of data samples. %The task of estimating the difficulty of data samples was addressed in different domain-specific ways, \eg~text length was used in natural language processing \cite{Tay-ACL-2019,Zhang-ISPASS-2021}, while the number or the size of objects were shown to work well in computer vision \cite{Shi-ECCV-2016,Soviany-CVIU-2021}. 
Despite having many successful applications \citep{Soviany-IJCV-2022,Wang-PAMI-2021}, there is no universal way to determine the difficulty of the data samples, making the data-level curriculum less applicable to scenarios where the difficulty is hard to estimate, \eg~classification of radar signals. The task-level and objective-level curriculum learning strategies suffer from similar issues, \eg~it is hard to create a curriculum when the model has to learn an easy task (binary classification) or the objective function is already convex.

\begin{figure}[!t]
\begin{center}
\centerline{\includegraphics[width=1\linewidth]{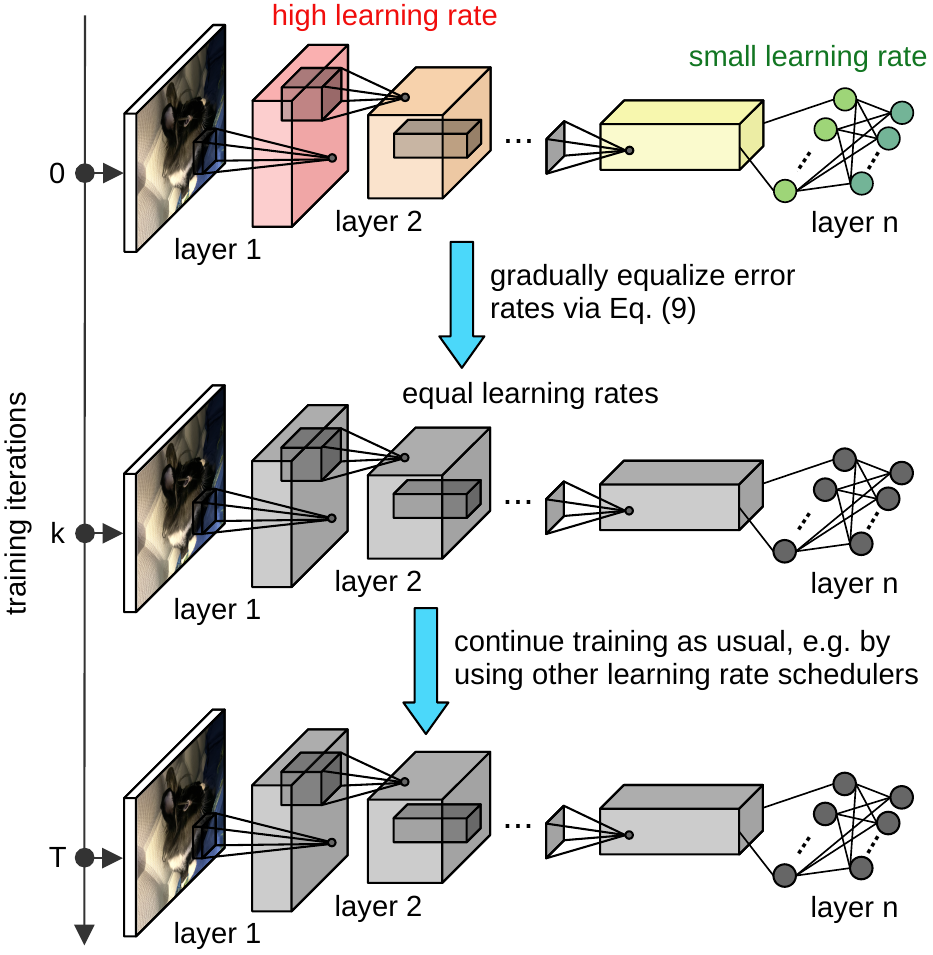}}
% \vspace{-0.25cm}
\caption{Training based on Learning Rate Curriculum.}
\label{fig_lerac}
\vspace{-0.5cm}
%\vspace{-0.9cm}
\end{center}
\end{figure}

Considering the above observations, we recognize the potential of model-level curriculum learning strategies of being applicable across a wider range of domains and tasks. To date, there are only a few works \citep{Burduja-ICIP-2021,Karras-ICLR-2018,Sinha-NIPS-2020} in the category of pure model-level curriculum learning methods. However, these methods have some drawbacks caused by their domain-dependent or architecture-specific design. % For instance, Karras \etal~\cite{Karras-ICLR-2018} gradually increase the resolution of input images as new layers are appended to a generative network, but the notion of input resolution does not exist in other domains, \eg~text. Burduja \etal~\cite{Burduja-ICIP-2021} blur the input images with Gaussian kernels, but this method is not applicable to an input format for which there is no convolution operation, \eg~tabular data. Sinha \etal~\cite{Sinha-NIPS-2020} apply Gaussian kernel smoothing on convolutional activation maps, but this operation makes less sense for a feed-forward neural network formed only of dense layers.
To benefit from the full potential of the model-level curriculum learning category, we propose LeRaC (\textbf{Le}arning \textbf{Ra}te \textbf{C}urriculum), a novel and simple curriculum learning approach which leverages the use of a different learning rate for each layer of a neural network to create a data-agnostic curriculum during the initial training epochs. More specifically, LeRaC assigns higher learning rates to neural layers closer to the input, gradually decreasing the learning rates as the layers are placed farther away from the input. This reduces the propagation of noise caused by the multiplication operations inside the network, a phenomenon that is more prevalent when the weights are randomly initialized. The learning rates increase at various paces during the first training iterations, until they all reach the same value, as illustrated in Figure~\ref{fig_lerac}. From this point on, the neural model is trained as usual. This creates a model-level curriculum learning strategy that is applicable to any domain and compatible with any neural network, generating higher performance levels regardless of the architecture, without adding any extra training time. To the best of our knowledge, we are the first to employ a different learning rate per layer to achieve the same effect as conventional (data-level) curriculum learning.

\begin{figure}[t]
  \centering
  \includegraphics[width=1.\linewidth]{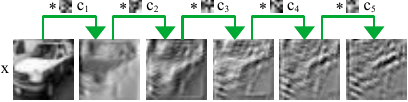}
  % \vspace{-0.25cm}
   \caption{Convolving an image of a car with random noise filters progressively increases the level of noise in the features. A theoretical proof of this observation is given in Appendix~\ref{app_proof}.}
   \label{fig-noise-conv}
   %\vspace{-0.4cm}
\end{figure}

As hinted above, the underlying hypothesis that justifies the use of LeRaC is that the level of noise grows from one neural layer to the next, especially when the input is multiplied with randomly initialized weights having low signal-to-noise ratios. We briefly illustrate this phenomenon through an example. Suppose an image $x$ is successively convolved with a set of random filters $c_1$, $c_2$, ..., $c_n$. Since the filters are uncorrelated, each filter distorts the image in a different way, degrading the information in $x$ with each convolution. The information in $x$ is gradually replaced by noise (see Fig.~\ref{fig-noise-conv}), \ie~the signal-to-noise ratio increases with each layer. Optimizing the filter $c_n$ to learn a pattern from the image convolved with $c_1$, $c_2$, ..., $c_{n-1}$ is suboptimal, because the filter $c_{n}$ will adapt to the noisy (biased) activation map induced by filters $c_1$, $c_2$, ..., $c_{n-1}$. This suggests that earlier filters need to be optimized sooner to reduce the level of noise of the activation map passed to layer $n$. In general, this phenomenon becomes more obvious as the layers get deeper, since the number of multiplication operations grows along the way. Hence, in the initial training stages, it makes sense to use gradually lower learning rates, as the layers get father away from the input. Our hypothesis is theoretically supported by Theorem \ref{prop_snr}, and empirically validated in Appendix~\ref{app_empirical_proof}.

% If we would employ the opposite strategy (anti-curriculum), training later layers at a faster pace, the later layers get adapted to the noise induced by earlier layers, which could likely lead to local optima or difficult training (due to the need of readapting to the earlier layers, once these layers start learning useful features).

% \vspace{-0.1cm}
We conduct comprehensive experiments on 12 data sets from the computer vision (CIFAR-10 \citep{Krizhevsky-TECHREP-2009}, CIFAR-100 \citep{Krizhevsky-TECHREP-2009}, Tiny ImageNet \citep{Russakovsky-IJCV-2015},
ImageNet-1K \citep{Russakovsky-IJCV-2015}, Food-101 \citep{Bossard-ECCV-2014}, UTKFace \citep{Zhang-CVPR-2017}, PASCAL VOC \citep{Everingham-IJCV-2010}), language (BoolQ \citep{Clark-NAACL-2019}, QNLI \citep{Wang-ICLR-2019},
RTE \citep{Wang-ICLR-2019}) and audio (ESC-50 \citep{Piczak-ACMMM-2015}, CREMA-D \citep{Cao-TAC-2014}) domains, considering various convolutional (ResNet-18 \citep{He-CVPR-2016}, Wide-ResNet-50 \citep{Zagoruyko-ArXiv-2016wide}, DenseNet-121 \citep{Gao-CVPR-2017}, YOLOv5 \citep{Jocher-zenodo-2022}), recurrent (LSTM \citep{Hochreiter-NC-1997}) and transformer (CvT \citep{Wu-ICCV-2021}, BERT \citep{Devlin-NAACL-2019}, SepTr \citep{Ristea-ARXIV-2022}) architectures. We compare our approach with the conventional training regime and Curriculum by Smoothing (CBS) \citep{Sinha-NIPS-2020}, our closest competitor. Unlike CBS, our performance improvements over the standard training regime are consistent across all data sets and models. Furthermore, we significantly surpass CBS in terms of training time, since there is no additional cost over the conventional training regime for LeRaC, whereas CBS adds Gaussian smoothing layers. We also compare with several data-level and task-level curriculum learning methods \citep{Dogan-ECCV-2020,Wang-ICCV-2023,Khan-arXiv-2024,Khan-BigComp-2023,Khan-RTIP-2023}, and show that our method scores best in most of the experiments.

In summary, our contribution is threefold:
\begin{itemize}
    \item We propose a novel and simple model-level curriculum learning strategy that creates a curriculum by updating the weights of each neural layer with a different learning rate, considering higher learning rates for the low-level feature layers and lower learning rates for the high-level feature layers.
    \item We empirically demonstrate the applicability to multiple domains (image, audio and text), the compatibility to several neural network architectures (convolutional neural networks, recurrent neural networks and transformers), and the time efficiency (no extra training time added) of LeRaC through a comprehensive set of experiments.
    \item We demonstrate our underlying hypothesis stating that the level of noise increases from one neural layer to another, both theoretically and empirically.
\end{itemize}

%\vspace{-0.2cm}
\section{Related Work}
%\vspace{-0.1cm}

\subsection{Curriculum Learning}

Curriculum learning was initially introduced by \citet{Bengio-ICML-2009} as a training strategy that helps machine learning models to generalize better when the training examples are presented in the ascending order of their difficulty. Extensive surveys on curriculum learning methods, including the most recent advancements on the topic, were conducted by \citet{Soviany-IJCV-2022} and \citet{Wang-PAMI-2021}. In the former survey, \citet{Soviany-IJCV-2022} emphasized that curriculum learning is not only applied at the data level, but also with respect to the other components involved in a machine learning approach, namely at the model level, the task level and the objective (loss) level. Regardless of the component on which curriculum learning is applied, the technique has demonstrated its effectiveness on a broad range of machine learning tasks, from computer vision \citep{Bengio-ICML-2009, Gui-FG-2017, Jiang-ICML-2018, Shi-ECCV-2016, Soviany-CVIU-2021, Chen-ICCV-2015, Sinha-NIPS-2020, Khan-arXiv-2024, Khan-BigComp-2023, Khan-RTIP-2023} to natural language processing \citep{Platanios-NAACL-2019, Kocmi-RANLP-2017, Spitkovsky-NIPS-2009, Liu-IJCAI-2018,Bengio-ICML-2009} and audio processing \citep{Ranjan-ACM-2018, Amodei-ICML-2016}.

The main challenge for the methods that build the curriculum at the data level is measuring the difficulty of the data samples, which is required to order the samples from easy to hard. Most studies have addressed the problem with human input \citep{Pentina-CVPR-2015, Sanchez-MICCAI-2019,Wei-WACV-2021} or metrics based on domain-specific heuristics. For instance, the text length \citep{Kocmi-RANLP-2017, Cirik-Arxiv-2016,Tay-ACL-2019,Zhang-ISPASS-2021} and the word frequency \citep{Bengio-ICML-2009, Liu-IJCAI-2018} have been employed in natural language processing. In computer vision, the samples containing fewer and larger objects have been considered to be easier in some works \citep{Soviany-CVIU-2021, Shi-ECCV-2016}. Other solutions employed difficulty estimators \citep{Ionescu-CVPR-2016} or even the confidence level of the predictions made by the neural network \citep{Gong-TIP-2016, Hacohen-Arxiv-2019} to approximate the complexity of the data samples. Other studies \citep{Khan-arXiv-2024, Khan-BigComp-2023, Khan-RTIP-2023} used the error of a previously trained model to estimate the difficulty of each sample.
Such solutions have shown their utility in specific application domains. Nonetheless, measuring the difficulty remains problematic when implementing standard (data-level) curriculum learning strategies, at least in some application domains. Therefore, several alternatives have emerged over time, handling the drawback and improving the conventional curriculum learning approach. In \citep{Kumar-ANIPS-2010}, the authors introduced self-paced learning to evaluate the learning progress when selecting training samples. The method was successfully employed in multiple settings \citep{Kumar-ANIPS-2010, Gong-TEVC-2019, Fan-AAAI-2017, Li-AAAI-2016, Jhou-PR-2018, Jiang-AAAI-2015, Ristea-INTERSPEECH-2021}. Furthermore, some studies combined self-paced learning with the traditional pre-computed difficulty metrics \citep{Jiang-AAAI-2015, Ma-ICML-2017}. An additional advancement related to self-paced learning is the approach called self-paced learning with diversity \citep{Jiang-NIPS-2014}. The authors demonstrated that enforcing a certain level of variety among the selected examples can improve the final performance. Another set of methods that bypass the need for predefined difficulty metrics is known as teacher-student curriculum learning \citep{Zhang-AS-2019, Wu-NIPS-2018}. In this setting, a teacher network learns a curriculum to supervise a student neural network.

Closer to our work, a few methods \citep{Karras-ICLR-2018,Sinha-NIPS-2020,Burduja-ICIP-2021} proposed to apply curriculum learning at the model level, by gradually increasing the learning capacity (complexity) of the neural architecture. Such curriculum learning strategies do not need to know the difficulty of the data samples, thus having a great potential to be useful in a broad range of tasks. For example, \citet{Karras-ICLR-2018} proposed to gradually add layers to generative adversarial networks during training, while increasing the resolution of the input images at the same time. They are thus able to generate realistic high-resolution images. However, their approach is not applicable to every domain, since there is no notion of resolution for some input data types, \eg~text. 
\citet{Sinha-NIPS-2020} presented a strategy that %does not build any curricula beforehand but executes CL implicitly via the training process. In the early phases of the training, they 
blurs the activation maps of the convolutional layers using Gaussian kernel layers, reducing the noisy information caused by the network initialization. The blur level is progressively reduced to zero by decreasing the standard deviation of the Gaussian kernels. With this mechanism, they obtain a training procedure that allows the neural network to see simple information at the start of the process and more intricate details towards the end. Curriculum by Smoothing (CBS) \citep{Sinha-NIPS-2020} was only shown to be useful for convolutional architectures applied in the image domain. Although we found that CBS is applicable to transformers by blurring the tokens, it is not necessarily applicable to any neural architecture, \eg~standard feed-forward neural networks. As an alternative to CBS, \citet{Burduja-ICIP-2021} proposed to apply the same smoothing process on the input image instead of the activation maps. The method was applied with success in medical image alignment. However, this approach is not applicable to natural language input, as it is not clear how to apply the blurring operation on the input text.

Different from \citet{Burduja-ICIP-2021} and \citet{Karras-ICLR-2018}, our approach is applicable to various domains, including but not limited to natural language processing, as demonstrated throughout our experiments. To the best of our knowledge, the only competing model-level curriculum method which is applicable to various domains is CBS \citep{Sinha-NIPS-2020}. Unlike CBS, LeRaC does not introduce new operations, such as smoothing with Gaussian kernels, during training. As such, our approach does not increase the training time with respect to the conventional training regime, as later shown in the experiments included in Section~\ref{sec_experiments}.

To classify our approach as a curriculum learning framework, we consider the extreme case when the learning rate is set to zero for later layers, which is equivalent to freezing those layers. This clearly reduces the learning capacity of the model. If layers are unfrozen one by one, the capacity of the model grows. LeRaC can be seen as a soft version of the model-level curriculum method described above. We thus classify LeRaC as a model-level curriculum method. However, our method can also be seen as a curriculum learning strategy that simplifies the optimization \citep{Pentina-CVPR-2015, Sanchez-MICCAI-2019,Wei-WACV-2021, Kocmi-RANLP-2017, Cirik-Arxiv-2016,Tay-ACL-2019,Zhang-ISPASS-2021, Bengio-ICML-2009, Liu-IJCAI-2018} in the early training stages by restricting the model updates (in a soft manner) to certain directions (corresponding to the weights of the earlier layers). Due to the imposed soft restrictions (lower learning rates for deeper layers), the optimization is easier at the beginning. As the training progresses, all directions become equally important, and the network is permitted to optimize the loss function in any direction. As the number of directions grows, the optimization task becomes more complex (it is harder to find the optimum). Hence, a relationship to curriculum learning can be discovered by noting that the complexity of the optimization increases over time, just as in curriculum learning.

In summary, we consider that the simplicity of our approach comes with many important advantages: applicability to any domain and task, compatibility with any neural network architecture, and time efficiency (adds no extra training time). We support all these claims through the comprehensive experiments presented in Section~\ref{sec_experiments}. 

\subsection{Learning Rate Schedulers}
There are some contributions \citep{Singh-ICMLA-2015,You-arXiv-2017} showing that using adaptive learning rates can lead to improved results. We explain how our method is different below. In \citep{Singh-ICMLA-2015}, the main goal is increasing the learning rate of certain layers as necessary, to escape saddle points. Different from \citet{Singh-ICMLA-2015}, our strategy reduces the learning rates of deeper layers, introducing soft optimization restrictions in the initial training epochs. \citet{You-arXiv-2017} proposed to train models with very large batches using a learning rate for each layer, by scaling the learning rate with respect to the norms of the gradients. The goal of \citet{You-arXiv-2017} is to specifically learn models with large batch sizes, \eg~formed of 8K samples. Unlike \citet{You-arXiv-2017}, we propose a more generic approach that can be applied to multiple architectures (convolutional, recurrent, transformer) under unrestricted training settings. 

\citet{Gotmare-ICLR-2019} point out that \emph{learning rate with warm-up and restarts} is an effective strategy to improve stability of training neural models using large batches. Different from LeRaC, this approach does not employ a different learning rate for each layer. Moreover, the strategy restarts the learning rate at different moments during the entire training process, while LeRaC is applied only during the first few training epochs. %Aside from these technical differences, our experiments already include a direct comparison of the two strategies for the CvT architecture, \ie~the baseline CvT uses warm-up and restarts. The results show that introducing LeRaC brings consistent improvements. We thus conclude that our strategy is a viable and distinct alternative to learning rate with warm-up and restarts. 

\subsection{Optimizers}
We consider Adam \citep{Kingma-ICLR-1015} and related optimizers as orthogonal approaches that perform the optimization rather than setting the learning rate. Our approach, LeRaC, only aims to guide the optimization during the initial training iterations by reducing the relevance of optimizing deeper network layers. Most of the baseline architectures used in our experiments are already based on Adam or some of its variations, \eg~AdaMax, AdamW \citep{Loshchilov-ICLR-2019}. LeRaC is applied in conjunction with these optimizers, showing improved performance over various architectures and application domains. This supports our claim that LeRaC is an orthogonal contribution to the family of Adam optimizers. 

%In the supplementary, we explain how LeRaC is different from learning rate schedulers and optimizers. We also present additional experiments to support our claims.
 
\section{Method}

% {\bf Preliminaries and notations.}
Deep neural networks are commonly trained on a set of labeled data samples denoted as:
\begin{equation}
S\!=\!\{(x_i, y_i) | x_i\!\in\!X, y_i\!\in\!Y, \forall i \in \{1,2,...,m \} \},
\end{equation}
where $m$ is the number of examples, $x_i$ is a data sample and $y_i$ is the associated label. The training process of a neural network $f$ with parameters $\theta$ consists of minimizing some objective (loss) function $\mathcal{L}$ that quantifies the differences between the ground-truth labels and the predictions of the model $f$:
\begin{equation}
\min_{\theta} \frac{1}{m} \sum_{i=1}^m \mathcal{L}\left(y_i, f(x_i, \theta) \right).
\end{equation}

The optimization is generally performed by some variant of Stochastic Gradient Descent (SGD), where the gradients are back-propagated from the neural layers closer to the output towards the neural layers closer to input through the chain rule. Let $f_1$, $f_2$, ...., $f_n$ and $\theta_1$, $\theta_2$, ..., $\theta_n$ denote the neural layers and the corresponding weights of the model $f$, such that the weights $\theta_j$ belong to the layer $f_j$, $\forall j \in \{1, 2, ...,n\}$. The output of the neural network for some training data sample $x_i \in X$ is formally computed as follows:
\begin{equation}
\hat{y}_i\!=\!f (x_i, \theta)\!=\!f_n\!\left(... f_2 \left( f_1 \left(x_i, \theta_1 \right), \theta_2 \right) ...., \theta_n \right)\!.
\end{equation}

To optimize the model via SGD, the weights are updated as follows:
\begin{equation}
\theta_j^{(t+1)} = \theta_j^{(t)} - \eta^{(t)} \cdot \frac{\partial \mathcal{L}}{\partial \theta_j^{(t)}}, \forall j \in \{1, 2, ...,n\},
\end{equation}
where $t$ is the index of the current training iteration, $\eta^{(t)} > 0$ is the learning rate at iteration $t$, and the gradient of $\mathcal{L}$ with respect to $\theta_j^{(t)}$ is computed via the chain rule. Before starting the training process, the weights $\theta_j^{(0)}$ are commonly initialized with random values, \eg~using Glorot initialization \citep{Glorot-AISTATS-2010}. 

% {\bf Learning Rate Curriculum.}

\citet{Sinha-NIPS-2020} suggested that the random initialization of the weights produces a large amount of noise in the information propagated through the neural model during the early training iterations, which can negatively impact the learning process. Due to the feed-forward processing that involves several multiplication operations, we argue that the noise level grows with each neural layer, from $f_j$ to $f_{j+1}$. This statement is confirmed by the following theorem:
\vspace{0.1cm}
\begin{theorem}\label{prop_snr}
Let $s_1=u_1+z_1$ and $s_2=u_2+z_2$ be two signals, where $u_1$ and $u_2$ are the clean components, and $z_1$ and $z_2$ are the noise components. The signal-to-noise ratio of the product between the two signals is lower than the signal-to-noise ratios of the two signals, \ie:
\begin{equation}
\SNR(s_1\cdot s_2) \leq \SNR(s_i), \forall i \in \{1, 2\}.
\end{equation}
\end{theorem}
\begin{proof}
The proof is given in Appendix \ref{app_proof}.
\end{proof}

The same issue can occur if the weights are pre-trained on a distinct task, where the misalignment of the weights with a new task is likely higher for the high-level (specialized) feature layers. To alleviate this problem, we propose to introduce a curriculum learning strategy that assigns a different learning rate $\eta_j$ to each layer $f_j$, as follows:
\begin{equation}
\theta_j^{(t+1)} = \theta_j^{(t)} - \eta_j^{(t)} \cdot \frac{\partial \mathcal{L}}{\partial \theta_j^{(t)}}, \forall j \in \{1, 2, ...,n\},
\end{equation}
such that:
\begin{equation}\label{eq_init}
\eta^{(0)} \geq \eta_1^{(0)} \geq \eta_2^{(0)} \geq ... \geq \eta_n^{(0)},
\end{equation}
\begin{equation}\label{eq_iter_k}
\eta^{(k)} = \eta_1^{(k)} = \eta_2^{(k)} = ... = \eta_n^{(k)},
\end{equation}
where $\eta_j^{(0)}$ are the initial learning rates and $\eta_j^{(k)}$ are the updated learning rates at iteration $k$. The condition formulated in Eq.~(\ref{eq_init}) indicates that the initial learning rate $\eta_j^{(0)}$ of a neural layer $f_j$ gets lower as the level of the respective neural layer becomes higher (farther away from the input). With each training iteration $t \leq k$, the learning rates are gradually increased, until they become equal, according to Eq.~(\ref{eq_iter_k}). Thus, our curriculum learning strategy is only applied during the early training iterations, where the noise caused by the misfit (randomly initialized or pre-trained) weights is most prevalent. Hence, $k$ is a hyperparameter of LeRaC that is usually adjusted such that $k\ll T$, where $T$ is the total number of training iterations. %In practice, we obtain optimal results by running LeRaC up to any epoch between $2$ and $7$.

At this point, various schedulers can be used to increase each learning rate $\eta_j$ from iteration $0$ to iteration $k$. We empirically observed that an exponential scheduler is a better option than linear or logarithmic schedulers. We thus propose to employ the exponential scheduler, which is based on the following rule:
\begin{equation}\label{eq_update_exp}
\eta_j^{(l)}\!=\!\eta_j^{(0)}\!\cdot\!c^{\frac{l}{k} \cdot \left( \log_c \eta_j^{(k)} - \log_c \eta_j^{(0)} \right)}\!, \forall l\!\in\!\{0,1,...,k \}.
\end{equation}
We set $c=10$ in Eq.~(\ref{eq_update_exp}) across all our experiments. This is because learning rates are usually expressed as a power of $c=10$, \eg~$10^{-4}$. If we start with a learning rate of $\eta_j^{(0)}=10^{-8}$ for some layer $j$ and we want to increase it to $\eta_j^{(k)}=10^{-4}$ during the first 5 epochs ($k=4$), the intermediate learning rates generated via Eq.~(\ref{eq_update_exp}) are $\eta_j^{(1)}\!=\!10^{-7}$, $\eta_j^{(2)}\!=\!10^{-6}$, $\eta_j^{(3)}\!=\!10^{-5}$ and $\eta_j^{(4)}\!=\!10^{-4}$. We thus believe it is more intuitive to understand what happens when setting $c=10$ in Eq.~(\ref{eq_update_exp}), as opposed to using some tuned value for $c$. To this end, we refrain from tuning $c$ and fix it to $c=10$.

In practice, we obtain optimal results by initializing the lowest learning rate $\eta_n^{(0)}$ with a value that is around five or six orders of magnitude lower than $\eta^{(0)}$, while the highest learning rate $\eta_1^{(0)}$ is always equal to $\eta^{(0)}$. Apart from such general practical notes, the exact LeRaC configuration for each neural architecture is established by tuning its two hyperparameters ($k$, $\eta_n^{(0)}$) on the available validation sets.

We underline that the output feature maps of a layer $j$ are affected $(i)$ by the misfit weights $\theta_j^{(0)}$ of the respective layer, and $(ii)$ by the input feature maps, which are in turn affected by the misfit weights of the previous layers $\theta_1^{(0)}, ..., \theta_{j-1}^{(0)}$. Hence, the noise affecting the feature maps increases with each layer processing the feature maps, being multiplied with the weights from each layer along the way. Our curriculum learning strategy imposes the training of the earlier layers at a faster pace, transforming the noisy weights into discriminative patterns. As noise from the earlier layer weights is eliminated, we train the later layers at faster and faster paces, until all learning rates become equal at epoch $k$.

% In the supplementary, we provide a detailed discussion about the intuition behind LeRaC, while also motivating why it works.

From a technical point of view, we note that our approach can also be regarded as a way to guide the optimization, which we see as an alternative to loss function smoothing. The link between curriculum learning and loss smoothing is discussed by \citet{Soviany-IJCV-2022}, who suggest that curriculum learning strategies induce a smoothing of the loss function, where the smoothing is higher during the early training iterations (simplifying the optimization) and lower to non-existent during the late training iterations (restoring the complexity of the loss function). LeRaC is aimed at producing a similar effect, but in a softer manner by dampening the importance of optimizing the weights of high-level layers in the early training iterations. Additionally, we empirically observe (see results in Appendix~\ref{app_empirical_proof}) that LeRaC tends to balance the training pace of low-level and high-level features, while the conventional regime seems to update the high-level layers at a faster pace. This could provide an additional intuitive explanation of why our method works better.

\section{Experiments}
\label{sec_experiments}

\subsection{Data Sets}
We perform experiments on 12 benchmarks: CIFAR-10 \citep{Krizhevsky-TECHREP-2009}, CIFAR-100 \citep{Krizhevsky-TECHREP-2009}, Tiny ImageNet \citep{Russakovsky-IJCV-2015}, ImageNet-1K \citep{Russakovsky-IJCV-2015}, Food-101 \citep{Bossard-ECCV-2014}, UTKFace \citep{Zhang-CVPR-2017}, PASCAL VOC 2007+2012 \citep{Everingham-IJCV-2010}, BoolQ \citep{Clark-NAACL-2019}, QNLI \citep{Wang-ICLR-2019}, RTE \citep{Wang-ICLR-2019}, CREMA-D \citep{Cao-TAC-2014}, and ESC-50 \citep{Piczak-ACMMM-2015}. We adopt the official data splits for the 12 benchmarks considered in our experiments. When a validation set is not available, we keep $10\%$ of the training data for validation. %Additional details about the data sets are provided in the supplementary.

%\subsection{Data Set Descriptions}

\noindent{\bf CIFAR-10.}
CIFAR-10 \citep{Krizhevsky-TECHREP-2009} is a popular data set for object recognition in images. It consists of 60,000 color images with a resolution of $32 \times 32$ pixels. An image depicts one of 10 object classes, each class having 6,000 examples. We use the official data split with a training set of 50,000 images and a test set of 10,000 images. 

\noindent{\bf CIFAR-100.}
The CIFAR-100 \citep{Krizhevsky-TECHREP-2009} data set is similar to CIFAR-10, except that it has 100 classes with 600 images per class. There are 50,000 training images and 10,000 test images.

\noindent{\bf Tiny ImageNet.}
Tiny ImageNet is a subset of ImageNet-1K \citep{Russakovsky-IJCV-2015} which provides 100,000 training images, 25,000 validation images and 25,000 test images representing objects from 200 different classes. The size of each image is $64 \times 64$ pixels.

\noindent{\bf ImageNet.}
ImageNet-1K \citep{Russakovsky-IJCV-2015} is the most popular bemchmark in computer vision, comprising about 1.2 million images from 1,000 object categories. We set the resolution of all images to $224 \times 224$ pixels.

\noindent{\bf Food-101.} Food-101 \cite{Bossard-ECCV-2014} is a data set that contains images from 101 food categories. For each category, there are 750 training images and 250 test images. Thus, the total number of images is 101,000. We resize all images to $224 \times 224$ pixels.
The test set is manually cleaned, while the training set is purposely left uncurated, being affected by labeling noise. This makes Food-101 suitable for testing the robustness of models to labeling noise. 

\noindent{\bf UTKFace.} The UTKFace data set \citep{Zhang-CVPR-2017} contains face images representing various gender, age and ethnic groups. It consists of 23,709 images of $200 \times 200$ pixels. The data set is divided into 16,597 training images, 3,556 validation images, and 3,556 test images. Each image is annotated with the corresponding age and gender label, which makes UTKFace suitable for evaluating models in a multi-task learning setup.

\noindent{\bf PASCAL VOC 2007+2012.} One of the most popular benchmarks for object detection is PASCAL VOC \citep{Everingham-IJCV-2010}. The data set consists of 21,503 images which are annotated with bounding boxes for 20 object categories. The official split has 16,551 training images and 4,952 test images. 

\noindent{\bf BoolQ.}
BoolQ \citep{Clark-NAACL-2019} is a question answering data set for yes/no questions containing 15,942 examples. The questions are naturally occurring, being generated in unprompted and unconstrained settings. Each example is a triplet of the form: \{question, passage, answer\}. %, with the title of the page as optional additional context. %The questions are gathered from anonymized, aggregated queries to the Google Search engine. 
We use the data split provided in the SuperGLUE benchmark \citep{Wang-NIPS-2019}, containing 9,427 examples for training, 3,270 for validation and 3,245 for testing.

\begin{table*}[!t]
    \caption{Optimal hyperparameter settings for the various neural architectures used in our experiments. Notice that $\eta_1^{(0)}$ is always equal to $\eta^{(0)}$, being set without tuning. This means that LeRaC has only two tunable hyperparameters, $k$ and $\eta_n^{(0)}$, while CBS \citep{Sinha-NIPS-2020} has three.}
    \label{tab_parameters}
\small{
  \begin{center}
  \setlength\tabcolsep{3.8pt}
  \renewcommand{\arraystretch}{1.1}
  \begin{tabular}{|l|l|c|c|c||c|c|c||c|c|}
    \hline
    \multirow{2}{*}{Model}    & \multirow{2}{*}{Optimizer}     & \multirow{2}{*}{Mini-batch}          &  \multirow{2}{*}{\#Epochs}      &  \multirow{2}{*}{$\eta^{(0)}$}  & \multicolumn{3}{c||}{CBS} & \multicolumn{2}{c|}{LeRaC}          \\
    \cline{6-10}
    & & & & & $\sigma$ & $d$ & $u$ & $k$ & $\eta_1^{(0)}$ - $\eta_n^{(0)}$\\
    \hline
    \hline
    ResNet-18       & SGD          &  64 &  100-200 & $10^{-1}$ & 1 & 0.9 & 2-5 & 5-7 & $10^{-1}$ - $10^{-8}$\\
    Wide-ResNet-50  & SGD          & 64 &   100-200 & $10^{-1}$ & 1 & 0.9 & 2-5  & 5-7 & $10^{-1}$ - $10^{-8}$ \\
    CvT-13          & AdaMax       & 64-128 & 150-200 & $2\!\cdot\!10^{-3}$ & 1 & 0.9 & 2-5 & 2-5 & $2\!\cdot\!10^{-3}$ - $2\!\cdot\!10^{-8}$ \\
        CvT-13$_{\mbox{\scriptsize{pre-trained}}}$ & AdaMax    & 64-128 & 25 & $5\!\cdot\!10^{-4}$ & 1 & 0.9 & 2-5 & 3-6 & $5\!\cdot\!10^{-4}$ - $5\!\cdot\!10^{-10}$\\
    \hline
    YOLOv5$_{\mbox{\scriptsize{pre-trained}}}$ & SGD    & 16 & 100 & $10^{-2}$ & 1 & 0.9 & 2 & 3 & $10^{-2}$ - $10^{-5}$\\
    \hline
    BERT$_{\mbox{\scriptsize{large-uncased}}}$           & AdaMax        & 10     & 7-25 & $5\!\cdot\!10^{-5}$ & 1 & 0.9 & 1 & 3 & $5\!\cdot\!10^{-5}$ - $5\!\cdot\!10^{-8}$\\
    LSTM           & AdamW         & 256-512 & 25-70 & $10^{-3}$ & 1 & 0.9 & 2 & 3-4 & $10^{-3}$ - $10^{-7}$ \\
    \hline    
    SepTR           & Adam         &  2  &  50  &  $10^{-4}$ & 0.8 & 0.9 & 1-3 & 2-5 & $10^{-4}$ - $10^{-8}$\\

    DenseNet-121   & Adam         &  64 &  50  &  $10^{-4}$ &  0.8 & 0.9 & 1-3  & 2-5 & $10^{-4}$ - $5\!\cdot\!10^{-8}$\\
    \hline
  \end{tabular}
  \end{center}
  }
  %\vspace{-0.3cm}
\end{table*}

\noindent{\bf QNLI.}
The QNLI (Question-answering Natural Language Inference) data set \citep{Wang-ICLR-2019} is a natural language inference benchmark automatically derived from SQuAD \citep{Rajpurkar-EMNLP-2016}. The data set contains \{question, sentence\} pairs and the task is to determine whether the context sentence contains the answer to the question. The data set is constructed on top of Wikipedia documents, each document being accompanied, on average, by 4 questions. We consider the data split provided in the GLUE benchmark \citep{Wang-ICLR-2019}, which comprises 104,743 examples for training, 5,463 for validation and 5,463 for testing.

\noindent{\bf RTE.} 
Recognizing Textual Entailment (RTE) \citep{Wang-ICLR-2019} is a natural language inference data set containing pairs of sentences with the target label indicating if the meaning of one sentence can be inferred from the other. The training subset includes 2,490 samples, the validation set 277 samples, and the test set 3,000 samples.

\noindent{\bf CREMA-D.}
The CREMA-D multi-modal database \citep{Cao-TAC-2014} is formed of 7,442 videos of 91 actors (48 male and 43 female) of different ethnic groups. The actors perform various emotions while uttering 12 particular sentences that evoke one of the 6 emotion categories: anger, disgust, fear, happy, neutral, and sad. Following previous work \citep{Ristea-INTERSPEECH-2021}, we conduct experiments only on the audio modality, dividing the set of audio samples into $70\%$ for training, $15\%$ for validation and $15\%$ for testing.

\noindent{\bf ESC-50.}
The ESC-50 \citep{Piczak-ACMMM-2015} data set is a collection of 2,000 samples of 5 seconds each, comprising 50 classes of various common sound events. Samples are recorded at a 44.1 kHz sampling frequency, with a single channel. In our evaluation, we employ the 5-fold cross-validation procedure, as described in related works \citep{Piczak-ACMMM-2015,Ristea-ARXIV-2022}.
% \vspace{-0.1cm}

\subsection{Experimental Setup}

\noindent{\bf Architectures.} To demonstrate the compatibility of LeRaC with multiple neural architectures, we select several convolutional, recurrent and transformer models. As representative convolutional neural networks (CNNs), we opt for ResNet-18 \citep{He-CVPR-2016}, Wide-ResNet-50 \citep{Zagoruyko-ArXiv-2016wide} and DenseNet-121 \citep{Gao-CVPR-2017}. For the object detection experiments on PASCAL VOC, we use the YOLOv5 \citep{Jocher-zenodo-2022} model based on the CSPDarknet53 \citep{Wang-CVPRW-2020} backbone, which is pre-trained on the MS COCO data set \citep{Lin-ECCV-2014}. As representative transformers, we consider CvT-13 \citep{Wu-ICCV-2021}, BERT$_{\mbox{\scriptsize{uncased-large}}}$ \citep{Devlin-NAACL-2019} and SepTr \citep{Ristea-ARXIV-2022}. For CvT, we consider both pre-trained and randomly initialized versions. We use an uncased large pre-trained version of BERT. As \citet{Ristea-ARXIV-2022}, we train SepTr from scratch.
In addition, we employ a long short-term memory (LSTM) network \citep{Hochreiter-NC-1997} to represent recurrent neural networks (RNNs). The recurrent neural network contains two LSTM layers, each having a hidden dimension of 256 components. These layers are preceded by one embedding layer with the embedding size set to 128 elements. The output of the last recurrent layer is passed to a classifier composed of two fully connected layers. The LSTM is activated by rectified linear units (ReLU). 
We apply the aforementioned models on distinct input data types, considering the intended application domain of each model. %\footnote{The only exception is DenseNet-121, which is applied on audio data.}. 
Hence, ResNet-18, Wide-ResNet-50, CvT and YOLOv5 are applied on images, BERT and LSTM are applied on text, and SepTr and DenseNet-121 are applied on audio. % For each domain, we perform custom preprocessing steps, which we detail in the supplementary.

\noindent{\bf Multi-task architectures.} To determine the impact of LeRaC on multi-task learning models, we conduct experiments on the UTKFace data set, where the face images are annotated with gender and age labels. We consider two models for the multi-task learning setup, namely ResNet-18 and CvT-13. Each model is jointly trained on the two tasks (gender prediction and age estimation). To each model, we attach two heads, one for gender classification and one for age estimation, respectively. The classification head is trained using the cross-entropy loss with respect to the gender label, while the regression head uses the mean squared error with respect to the age label. The models are trained using a joint objective defined as follows:
\begin{equation}
\mathcal{L}_{\mbox{\tiny{MTL}}} \!=\! \frac{1}{m} \sum_{i=1}^m \mathcal{L}_{\mbox{\tiny{CE}}} \left(y^{g}_i, \hat{y}^{g}_i \right) \!+\! \lambda\!\cdot\!\mathcal{L}_{\mbox{\tiny{MSE}}} \left(y^{a}_i, \hat{y}^{a}_i \right),
\end{equation}
where $y^{g}_i$ and $y^{a}_i$ are the ground-truth gender and age labels, $\hat{y}^{g}_i$ and $\hat{y}^{a}_i$ are the predicted gender and age labels, $\lambda \in \mathbb{R}^+$ is a weight factor, and $\mathcal{L}_{\mbox{\tiny{CE}}}$ is the cross-entropy loss for the gender prediction task, defined as:
\begin{equation}
\mathcal{L}_{\mbox{\tiny{CE}}}\! \left(y^{g}_i, \hat{y}^{g}_i \right)\!=\!-\left( y^g_i \log(\hat{y}^g_i)\!+\!(1\!-\!y^g_i) \log(1 - \hat{y}^g_i) \right),
\end{equation}
and $\mathcal{L}_{\mbox{\tiny{MSE}}}$ is the mean squared error for the age estimation task, defined as:
\begin{equation}
\mathcal{L}_{\mbox{\tiny{MSE}}} \left(y^{a}_i, \hat{y}^{a}_i \right) = (y^{a}_i - \hat{y}^{a}_i)^2.
\end{equation}
The factor $\lambda$ ensures the two tasks are equally important by weighting $\mathcal{L}_{\mbox{\tiny{MSE}}}$ to have approximately the same range of values as $\mathcal{L}_{\mbox{\tiny{CE}}}$. As such, we set $\lambda=10$.

% \vspace{-0.1cm}
\noindent{\bf Baselines.} 
We compare LeRaC with two baselines: the conventional training regime (which uses early stopping, reduces the learning rate on plateau, and employs linear warm-up and cosine annealing when required) and the state-of-the-art Curriculum by Smoothing~\citep{Sinha-NIPS-2020}. For CBS, we use the official code released by \citet{Sinha-NIPS-2020} at \url{https://github.com/pairlab/CBS}, to ensure the reproducibility of their method in our experimental settings, which include a more diverse selection of input data types and neural architectures. In addition, we compare with several data-level and task-level curriculum learning methods \citep{Dogan-ECCV-2020,Wang-ICCV-2023,Khan-arXiv-2024,Khan-BigComp-2023,Khan-RTIP-2023} on CIFAR-10 and CIFAR-100.

To apply CBS to non-convolutional architectures, we use 1D convolutional layers based on Gaussian filters with a receptive field of 3. For transformers, we integrate a 1D Gaussian layer before each transformer block, so the smoothing is applied on the sequence of tokens. Similarly, for recurrent neural networks, before each LSTM layer, we process the sequence of tokens with 1D convolutional layers based on Gaussian filters. For both transformers and RNNs, we anneal, during training, the standard deviation of the Gaussian filters to enhance the information propagated through the network. This approach mirrors the implementation of CBS for convolutional neural networks.

% \vspace{-0.1cm}
\noindent{\bf Hyperparameter tuning.} 
We tune all hyperparameters on the validation set of each benchmark. In Table \ref{tab_parameters}, we present the optimal hyperparameters chosen for each architecture. In addition to the standard parameters of the training process, we report the parameters that are specific for the CBS \citep{Sinha-NIPS-2020} and LeRaC strategies. In the case of CBS, $\sigma$ denotes the standard deviation of the Gaussian kernel, $d$ is the decay rate for $\sigma$, and $u$ is the decay step. Regarding the parameters of LeRaC, $k$ represents the number of iterations used in Eq.~(\ref{eq_update_exp}), and $\eta_1^{(0)}$ and $\eta_n^{(0)}$ are the initial learning rates for the first and last layers of the architecture, respectively. We set $\eta_1^{(0)} = \eta^{(0)}$ and $c= 10$ in all experiments, without tuning. In addition, the intermediate learning rates $\eta_j^{(0)}$, $\forall j \in \{2, 3, ...,n-1\}$, are automatically set to be equally distanced between $\eta_1^{(0)}$ and $\eta_n^{(0)}$. Moreover, $\eta_j^{(k)} = \eta^{(0)}$, \ie~the initial learning rates of LeRaC converge to the original learning rate set for the conventional training regime. All models are trained with early stopping and the learning rate is reduced by a factor of $10$ when the loss reaches a plateau. We use linear warm-up with cosine annealing, whenever it is found useful for models based on conventional or CBS training. The learning rate warm-up is switched off for LeRaC to avoid unwanted interactions with our training strategy.
Except for the pre-trained models, the weights of all models are initialized with Glorot initialization \citep{Glorot-AISTATS-2010}.

We underline that some parameters are the same across all data sets, while others need to be established per data set. For example, the parameter $u$ of CBS and the parameter $k$ of LeRaC are validated on each data set. As such, for the ResNet-18 model, the parameter $u$ of CBS takes one value on each data set (CIFAR-10, CIFAR-100, Tiny ImageNet, ImageNet, Food-101, UTKFace), but the values of $u$ on all five data sets can range between 2 and 5. Similarly, the parameter $k$ of LeRaC takes one value per data set, with the range of values being 5-7. In Table \ref{tab_parameters}, we aggregate the optimal parameters of each model for all data sets. This explains why some hyperparameters are specified in terms of ranges.

\noindent{\bf Setting the initial learning rates.}
We should emphasize that the different learning rates $\eta_j^{(0)}$, $\forall j \in \{1,2,...,n \}$, are not optimized nor tuned during training. Instead, we set the initial learning rates $\eta_j^{(0)}$ through validation, such that $\eta_n^{(0)}$ is around five or six orders of magnitude lower than $\eta^{(0)}$, and $\eta_1^{(0)}=\eta^{(0)}$. After initialization, we apply our exponential scheduler, until all learning rates become equal at iteration $k$. 
In addition, we would like to underline that the difference $\delta$ between the initial learning rates of consecutive layers is automatically set based on the range given by $\eta_1^{(0)}$ and $\eta_n^{(0)}$. For example, let us consider a network with 5 layers. If we choose $\eta_1^{(0)}=10^{-1}$ and $\eta_5^{(0)}=10^{-2}$, then the intermediate initial learning rates are automatically set to $\eta_2^{(0)}=10^{-1.25}$, $\eta_3^{(0)}=10^{-1.5}$, $\eta_4^{(0)}=10^{-1.75}$, \ie~$\delta$ is used in the exponent and is equal to $-0.25$ in this case. To obtain the intermediate learning rates according to this example, we actually apply the exponential scheduler defined in Eq.~(\ref{eq_update_exp}). This reduces the number of tunable hyperparameters from $n$ (the number layers) to two, namely $\eta_1^{(0)}$ and $\eta_n^{(0)}$. We go even further, setting $\eta_1^{(0)}=\eta^{(0)}$ without tuning, in all our experiments. Hence, tuning is only performed for the initial learning rate of the last layer, namely $\eta_n^{(0)}$. Although tuning all $\eta_j^{(0)}$, $\forall j \in \{1, 2, ...,n\}$, might lead to better results, we refrain from meticulously tuning every possible value to avoid overfitting in hyperparameter space.

\noindent{\bf Number of hyperparameters.} We further emphasize that LeRaC adds only two additional tunable hyperparameters with respect to the conventional training regime. These are the lowest learning rate $\eta_n^{(0)}$ and the number of iterations $k$ to employ LeRaC. We reduce the number of hyperparameters that require tuning by using a fixed rule to adjust the intermediate learning rates, \eg~by employing an exponential scheduler, or by fixing some hyperparameters, \eg~$c=10$. In contrast, CBS \citep{Sinha-NIPS-2020} has three additional hyperparameters, thus having one extra hyperparameter compared with LeRaC. Furthermore, we note that data-level curriculum methods also introduce additional hyperparameters. Even a simple method that splits the examples into easy-to-hard batches that are gradually added to the training set requires at least two hyperparameters: the number of batches, and the number of iterations before introducing a new training batch. We thus believe that, in terms of the number of additional hyperparameters, LeRaC is comparable to CBS and other curriculum learning strategies. We emphasize that the same happens if we look at new optimizers, \eg~Adam \citep{Kingma-ICLR-1015} adds three additional hyperparameters compared with SGD.

\begin{table*}[!t]
\caption{Average accuracy rates (in \%) over 5 runs on CIFAR-10, CIFAR-100 and Tiny ImageNet for various neural models based on different training regimes: learning rate decay, linear warm-up, cosine annealing, constant learning rate, and LeRaC. The accuracy of the best training regime in each experiment is highlighted in bold.}\label{tab_training_regime}
\small{
  \begin{center}
  \setlength\tabcolsep{0.14cm}
  \begin{tabular}{|l|l|c|c|c|c|c|}
    \hline
    Model       & Training Regime     & CIFAR-10  & CIFAR-100 & Tiny ImageNet \\
    \hline    
    \hline
    \multirow{3}{*}{ResNet-18}         & learning rate decay               &  $89.20 {\pm 0.43}$   &  $71.70 {\pm 0.06}$   & $57.41 {\pm 0.05}$ \\
    & constant learning rate & $72.30 {\pm 1.08}$ & $62.06 {\pm 0.41}$  & $49.42 {\pm 0.37}$ \\
             & LeRaC (ours)    &  $\mathbf{89.56} {\pm 0.16}$   &  $\mathbf{72.72} {\pm 0.12}$   & $\mathbf{57.86} {\pm 0.20}$ \\
    \hline   
    \multirow{3}{*}{Wide-ResNet-50}     & learning rate decay               &  $91.22 {\pm 0.24}$   &  $68.14 {\pm 0.16}$   & $55.97 {\pm 0.30}$ \\
    & constant learning rate & $86.62 {\pm 0.27}$ & $61.67 {\pm 0.12}$ & $41.87 {\pm 0.61}$ \\
         & LeRaC (ours)    &  $\mathbf{91.58} {\pm 0.16}$   &  $\mathbf{69.38} {\pm 0.26}$   & $\mathbf{56.48} {\pm 0.60}$ \\
    \hline    
    \multirow{3}{*}{CvT-13}             & linear warm-up + cosine annealing                &  $71.84 {\pm 0.37}$   &  $41.87 {\pm 0.16}$   & $33.38 {\pm 0.27}$ \\
    & constant learning rate & $71.75 {\pm 0.07}$ & $41.62 {\pm 0.20}$& $30.68 {\pm 0.10}$ \\
                 & LeRaC (ours)    &  $\mathbf{72.90} {\pm 0.28}$   &  $\mathbf{43.46} {\pm 0.18}$   & $\mathbf{33.95} {\pm 0.28}$ \\
    \hline
      \multirow{3}{*}{CvT-13$_{\mbox{\scriptsize{pre-trained}}}$} & cosine annealing              &  $93.06 {\pm 0.06}$   &  $77.76 {\pm 0.38}$   &  $70.91 {\pm 0.24}$\\
      & constant learning rate & $93.56 {\pm 0.05}$ & $77.80 {\pm 0.16}$ & $70.71 {\pm 0.35}$ \\
     & LeRaC (ours)    &  $\mathbf{94.15} {\pm 0.03}$   &  $\mathbf{78.93} {\pm 0.05}$   & $\mathbf{71.34} {\pm 0.08}$ \\
    \hline
  \end{tabular}
  \end{center}
}
    % \vspace{-0.3cm}
\end{table*}

\noindent{\bf Avoiding too large learning rates.}
In principle, a larger learning rate implies a larger update. However, if the learning rate is too high, the model can actually diverge. This is because the gradient describes the loss function in the vicinity of the current location, providing no guarantee for the value of the loss outside this vicinity. Our implementation takes this aspect into account. Instead of increasing the learning rate for earlier layers, we reduce the learning rate for the deeper layer to avoid divergence. More precisely, we set the learning rate for the first layer $\eta_1^{(0)}$ to the original learning rate $\eta^{(0)}$ and the other initial learning rates are gradually reduced with each layer. During training, the lower learning rates are gradually increased, until epoch $k$. Hence, LeRaC actually slows down the learning for deeper layers, until the earlier layers have learned representative features.

% \vspace{-0.1cm}
\noindent{\bf Evaluation.} For the classification tasks, we evaluate all models in terms of the accuracy rate. For the regression task (age estimation), we use the mean absolute error. For the object detection task, we employ the mean Average Precision (mAP) at an intersection over union (IoU) threshold of 0.5. We repeat the training process of each model for 5 times and report the average performance and the standard deviation.

\begin{table*}[!t]
\caption{Average accuracy rates (in \%) over 5 runs on CIFAR-10, CIFAR-100, Tiny ImageNet, ImageNet and Food-101 for various neural models based on different training regimes: conventional, CBS \citep{Sinha-NIPS-2020} and LeRaC. The accuracy of the best training regime in each experiment is highlighted in bold.}\label{tab_vision}
\small{
  \begin{center}
  \setlength\tabcolsep{0.14cm}
  \begin{tabular}{|l|l|c|c|c|c|c|}
    \hline
    Model       & Training Regime     & CIFAR-10  & CIFAR-100 & Tiny ImageNet & ImageNet & Food-101\\
    \hline    
    \hline
             & conventional               &  $89.20 {\pm 0.43}$   &  $71.70 {\pm 0.06}$   & $57.41 {\pm 0.05}$ & $68.44 {\pm 0.65}$ &  $68.31 {\pm 0.09}$\\
    ResNet-18          & CBS %\citep{Sinha-NIPS-2020}
    &  $89.53 {\pm 0.22}$   &  $\mathbf{72.80}
    {\pm 0.18}$   & $55.49 {\pm 0.20}$ & $71.02 {\pm 0.80}$ & $65.09 {\pm 0.47}$ \\
             & LeRaC (ours)    &  $\mathbf{89.56} {\pm 0.16}$   &  $72.72 {\pm 0.12}$   & $\mathbf{57.86} {\pm 0.20}$ & $ \mathbf{71.96} {\pm 0.72}$ & $\mathbf{69.57}  {\pm 0.07}$\\
    \hline   
         & conventional               &  $91.22 {\pm 0.24}$   &  $68.14 {\pm 0.16}$   & $55.97 {\pm 0.30}$ & $70.25 {\pm 0.82}$ & $67.54 {\pm 0.66}$\\
    Wide-ResNet-50     & CBS %\citep{Sinha-NIPS-2020}
    &  $89.05 {\pm 1.00}$   &  $65.73 {\pm 0.36}$   & $48.30 {\pm 1.53}$ & $72.10 {\pm 0.71}$ & $58.95 {\pm 1.80}$\\
         & LeRaC (ours)    &  $\mathbf{91.58} {\pm 0.16}$   &  $\mathbf{69.38} {\pm 0.26}$   & $\mathbf{56.48} {\pm 0.60}$ & $\mathbf{72.49} {\pm 0.64}$ & $\mathbf{67.96} {\pm 0.35}$\\
    \hline    
                 & conventional               &  $71.84 {\pm 0.37}$   &  $41.87 {\pm 0.16}$   & $33.38 {\pm 0.27}$ & $81.33 {\pm 0.75}$ & $39.17 {\pm 1.26}$\\
    CvT-13             & CBS %\citep{Sinha-NIPS-2020}
    &  $72.64 {\pm 0.29}$   &  $\mathbf{44.48} {\pm 0.40}$   & $33.56 {\pm 0.36}$ & $80.42 {\pm 0.58}$ &$38.63 {\pm 0.49}$ \\
                 & LeRaC (ours)    &  $\mathbf{72.90} {\pm 0.28}$   &  $43.46 {\pm 0.18}$   & $\mathbf{33.95} {\pm 0.28}$ & $\mathbf{82.19} {\pm 0.68}$ & $\mathbf{41.42} {\pm 0.72}$\\
    \hline
      & conventional               &  $93.56 {\pm 0.05}$   &  $77.80 {\pm 0.16}$   & $70.71 {\pm 0.35}$ & - & $85.22 {\pm 0.11}$ \\
    CvT-13$_{\mbox{\scriptsize{pre-trained}}}$  & CBS %\citep{Sinha-NIPS-2020}
    &  $85.85 {\pm 0.15}$   &  $62.35 {\pm 0.48}$   & $68.41 {\pm 0.13}$ & - & $81.41 {\pm 0.42}$\\
     & LeRaC (ours)    &  $\mathbf{94.15} {\pm 0.03}$   &  $\mathbf{78.93} {\pm 0.05}$   & $\mathbf{71.34} {\pm 0.08}$ & - & $\mathbf{86.05} {\pm 0.08}$ \\
    \hline
  \end{tabular}
  \end{center}
}
    % \vspace{-0.3cm}
\end{table*}

\subsection{Domain-Specific Preprocessing}

\noindent{\bf Image preprocessing.}
For the image classification experiments, we apply the same data preprocessing approach as \citet{Sinha-NIPS-2020}. Hence, we normalize the images and maintain their original resolution, $32 \times 32$ pixels for CIFAR-10 and CIFAR-100, $64 \times 64$ pixels for Tiny ImageNet, $224 \times 224$ pixels for ImageNet and Food-101, and $200 \times 200$ pixels for UTKFace. Similar to \citet{Sinha-NIPS-2020}, we do not employ data augmentation.

\noindent{\bf Text preprocessing.}
For the text classification experiments with BERT, we lowercase all words and add the classification token ([CLS]) at the start of the input sequence. We add the separator token ([SEP]) to delimit sentences. For the LSTM network, we lowercase all words and replace them with indexes from vocabularies constructed from the training set. The input sequence length is limited to $512$ tokens for BERT and $200$ tokens for LSTM.

\noindent{\bf Speech preprocessing.}
The speech preprocessing steps are carried out following \citet{Ristea-ARXIV-2022}. We thus transform each audio sample into a time-frequency matrix by computing the discrete Short Time Fourier Transform (STFT) with $N_x$ FFT points, using a Hamming window of length $L$ and a hop size $R$.
For CREMA-D, we first standardize all audio clips to a fixed dimension of $4$ seconds by padding or clipping the samples. Then, we apply the STFT with $N_x=1024$, $R=64$ and a window size of $L=512$. For ESC-50, we keep the same values for $N_x$ and $L$,
but we increase the hop size to $R = 128$. Next, for each STFT, we compute the square root of the magnitude and map the values to $128$ Mel bins. The result is converted to a logarithmic scale and normalized to the interval $[0, 1]$. Furthermore, in all our speech classification experiments, we use the following data augmentation methods: noise perturbation, time shifting, speed perturbation, mix-up and SpecAugment \citep{Park-INTERSPEECH-2019}.

\begin{table*}[t]
\caption{ Multi-task learning results for ResNet-18 and CvT-13 (pre-trained) on UTKFace, using three different training regimes: conventional, CBS \citep{Sinha-NIPS-2020} and LeRaC. We report the accuracy (in \%) for gender prediction and the mean absolute error (MAE) for age estimation. The $\downarrow$ and $\uparrow$ symbols indicate when lower or upper values are better, respectively. The best scores are highlighted in bold.}
% \vspace{-0.25cm}
\label{tab_multitask} % is used to refer this table in the text
\small{
\begin{center}
\begin{tabular}{|l| l | c | c |} 
\hline
 Model & Training Regime & Gender Accuracy $\uparrow$ & Age MAE $\downarrow$ \\
 \hline
 \hline
\multirow{3}{*}{ResNet-18} & conventional & $88.63 {\pm 0.12}$ & $6.75 {\pm 0.22}$ \\
& CBS & $89.23 {\pm 0.11}$ & $6.24 {\pm 0.22}$ \\
& LeRaC (ours) & $\mathbf{90.07} {\pm 0.12}$ & $\mathbf{5.97} {\pm 0.20}$ \\
\hline
\multirow{3}{*}{CvT-13$_{\mbox{\scriptsize{pre-trained}}}$} & conventional & $92.57 {\pm 0.15}$ & $4.78 {\pm 0.18}$ \\
& CBS & $92.61 {\pm 0.14}$ & $4.61 {\pm 0.17}$ \\
& LeRaC (ours) & $\mathbf{93.19} {\pm 0.14}$ & $\mathbf{4.06} {\pm 0.15}$ \\
\hline
\end{tabular}
\end{center}
}
\captionsetup{justification=centering}
\end{table*}

\begin{table*}[t]
\caption{Object detection results of YOLOv5 on PASCAL VOC, using three different training regimes: conventional, CBS \citep{Sinha-NIPS-2020} and LeRaC. The best mAP is highlighted in bold.}
% \vspace{-0.25cm}
\label{tab_obj_detection} % is used to refer this table in the text
\small{
\begin{center}
\begin{tabular}{|l|c | c | c |} 
\hline
Training Regime & conventional & CBS & LeRaC (ours) \\
 \hline
 \hline
mAP & $0.832 {\pm 0.006}$ & $0.829 {\pm 0.003}$ & $\mathbf{0.846} {\pm 0.004}$ \\
\hline
\end{tabular}
\end{center}
}
\captionsetup{justification=centering}

\end{table*}

\subsection{Preliminary Results}
\label{section_preresults}

We present preliminary experiments to show the effect of various learning rate schedulers for different architectures. For each architecture, we compare the constant learning rate scheduler with an adaptive learning rate scheduler. The aim is to find the best scheduler for the conventional training regime, which is used as baseline in the subsequent experiments. Table~\ref{tab_training_regime} showcases the preliminary results on CIFAR-10, CIFAR-100 and Tiny ImageNet. We compare the outcomes of the adaptive and constant learning rate schedulers with those of LeRaC. In most cases, the adaptive scheduler yields better results than the constant learning rate. Using a constant learning rate seems to work only for the pre-trained CvT-13. Notably, the analysis also reveals that LeRaC consistently outperforms the other baseline schedulers, achieving the highest accuracy rates across all data sets.

We emphasize that, for the subsequent experiments, the conventional regime is always represented by the best scheduler among the following options: learning rate decay, learning rate warm-up, cosine annealing, or combinations of the aforementioned options.

\begin{table*}[!t]
\caption{Left side: average accuracy rates (in \%) over 5 runs on BoolQ, RTE and QNLI for BERT and LSTM. Right side: average accuracy rates (in \%) over 5 runs on CREMA-D and ESC-50 for SepTr and DenseNet-121. In both domains (text and audio), the comparison is between different training regimes: conventional, CBS \citep{Sinha-NIPS-2020} and LeRaC. The accuracy of the best training regime in each experiment is highlighted in bold.}
  \label{tab_text}
\small{
  \setlength\tabcolsep{2.4pt}
  \begin{center}
  \begin{tabular}{|l||l|c|c|c||l|c|c|}
    \hline
    {Training}     & \multicolumn{4}{c||}{Text}    & \multicolumn{3}{c|}{Audio}      \\
    \cline{2-8}
   Regime & Model    &  BoolQ          & RTE                 &  QNLI    &   Model &  CREMA-D     & ESC-50      \\
    \hline    
    \hline
    conventional  &               &  $74.12 {\pm0.32}$   &  $74.48 {\pm1.36}$  &  $92.13 {\pm0.08}$ & & $70.47 {\pm0.67}$    & $91.13 {\pm0.33}$\\
   CBS %\citep{Sinha-NIPS-2020}
   & BERT$_{\mbox{\scriptsize{large}}}$  &   $74.37 {\pm1.11}$   &  $74.97 {\pm1.96}$  &  $91.47 {\pm0.22}$   & SepTr &  $69.98 {\pm0.71}$    & $91.15 {\pm0.41}$\\
     LeRaC (ours)   &  &  $\mathbf{75.55} {\pm0.66}$   &  $\mathbf{75.81} {\pm0.29}$  &  $\mathbf{92.45} {\pm0.13}$ &  & $\mathbf{70.95} {\pm0.56}$    & $\mathbf{91.58} {\pm0.28}$\\
    \hline  
     conventional           &               &  $64.40 {\pm1.37}$  &  $54.12 {\pm1.60}$  & $59.42 {\pm0.36}$ &  &  $67.21 {\pm0.12}$  & $88.91 {\pm0.11}$  \\
    CBS %\cite{Sinha-NIPS-2020}
    & LSTM            &  $64.75 {\pm1.54}$  &  $54.03 {\pm0.45}$  & $59.89 {\pm0.38}$   &DenseNet-121  & $68.16 {\pm0.19}$  & $88.76 {\pm0.17}$\\
       LeRaC (ours)         &    &  $\mathbf{65.80} {\pm0.33}$  &  $\mathbf{55.71} {\pm1.04}$  & $\mathbf{59.98} {\pm0.34}$ &  & $\mathbf{68.99} {\pm0.08}$  & $\mathbf{90.02} {\pm0.10}$ \\
    \hline
  \end{tabular}
  \end{center}
}
    % \vspace{-0.35cm}
    
\end{table*}

\subsection{Main Results}
\label{section_results}

\noindent{\bf Image classification.} In Table~\ref{tab_vision}, we present the image classification results on CIFAR-10, CIFAR-100, Tiny ImageNet, ImageNet and Food-101. Since CvT-13 is pre-trained on ImageNet, it does not make sense to fine-tune it on ImageNet. Thus, the respective results are not reported. On the one hand, there are two scenarios (ResNet-18 on CIFAR-100, and CvT-13 on CIFAR-100) in which CBS provides the largest improvements over the conventional regime, surpassing LeRaC in the respective cases. On the other hand, there are more than 10 scenarios where CBS degrades the accuracy with respect to the standard training regime. This shows that the improvements attained by CBS are inconsistent across models and data sets. Unlike CBS, our strategy surpasses the baseline regime in all 19 cases, thus being more consistent. In 8 of these cases, the accuracy gains of LeRaC are higher than $1\%$. Moreover, LeRaC outperforms CBS in 17 out of 19 cases. We thus consider that LeRaC can be regarded as a better choice than CBS, bringing consistent performance gains.

\noindent{\bf Multi-task learning.} In Table~\ref{tab_multitask}, we include the multi-task learning results on the UTKFace data set \citep{Zhang-CVPR-2017}. We evaluate the multi-task ResNet-18 and CvT-13$_{\mbox{\scriptsize{pre-trained}}}$ models under various training regimes, reporting the accuracy rates for gender prediction, and the mean absolute errors for age estimation, respectively. LeRaC achieves the best scores in each and every case, surpassing the other training regimes in the multi-task learning setup. Moreover, its results are statistically better with respect to both competing regimes. In contrast, the CBS regime remains in the statistical margin of the conventional regime for the pre-trained CvT-13 network.

% \vspace{-0.1cm}
\noindent{\bf Object detection.} In Table~\ref{tab_obj_detection}, we include the object detection results of YOLOv5 \citep{Jocher-zenodo-2022} based on different training regimes on PASCAL VOC 2007+2012 \citep{Everingham-IJCV-2010}. LeRaC exhibits a superior mAP score, significantly surpassing the other training regimes. In contrast, CBS leads to suboptimal performance, hinting towards the inconsistency of CBS across different evaluation scenarios.

% \vspace{-0.1cm}
\noindent{\bf Text classification.}
In Table~\ref{tab_text} (left side), we report the text classification results on BoolQ, RTE and QNLI. Here, there are two cases (BERT on QNLI and LSTM on RTE) where CBS leads to performance drops compared with the conventional training regime. In all other cases, the improvements of CBS are below $0.6\%$. Just as in the image classification experiments, LeRaC brings accuracy gains for each and every model and data set. In four out of six scenarios, the accuracy gains yielded by LeRaC are higher than $1.3\%$. Once again, LeRaC proves to be the most consistent regime, generally surpassing CBS by significant margins.

\begin{figure*}[!t]
\begin{subfigure}{.5\textwidth}
  \centering
  % include third image
  \includegraphics[width=.92\linewidth]{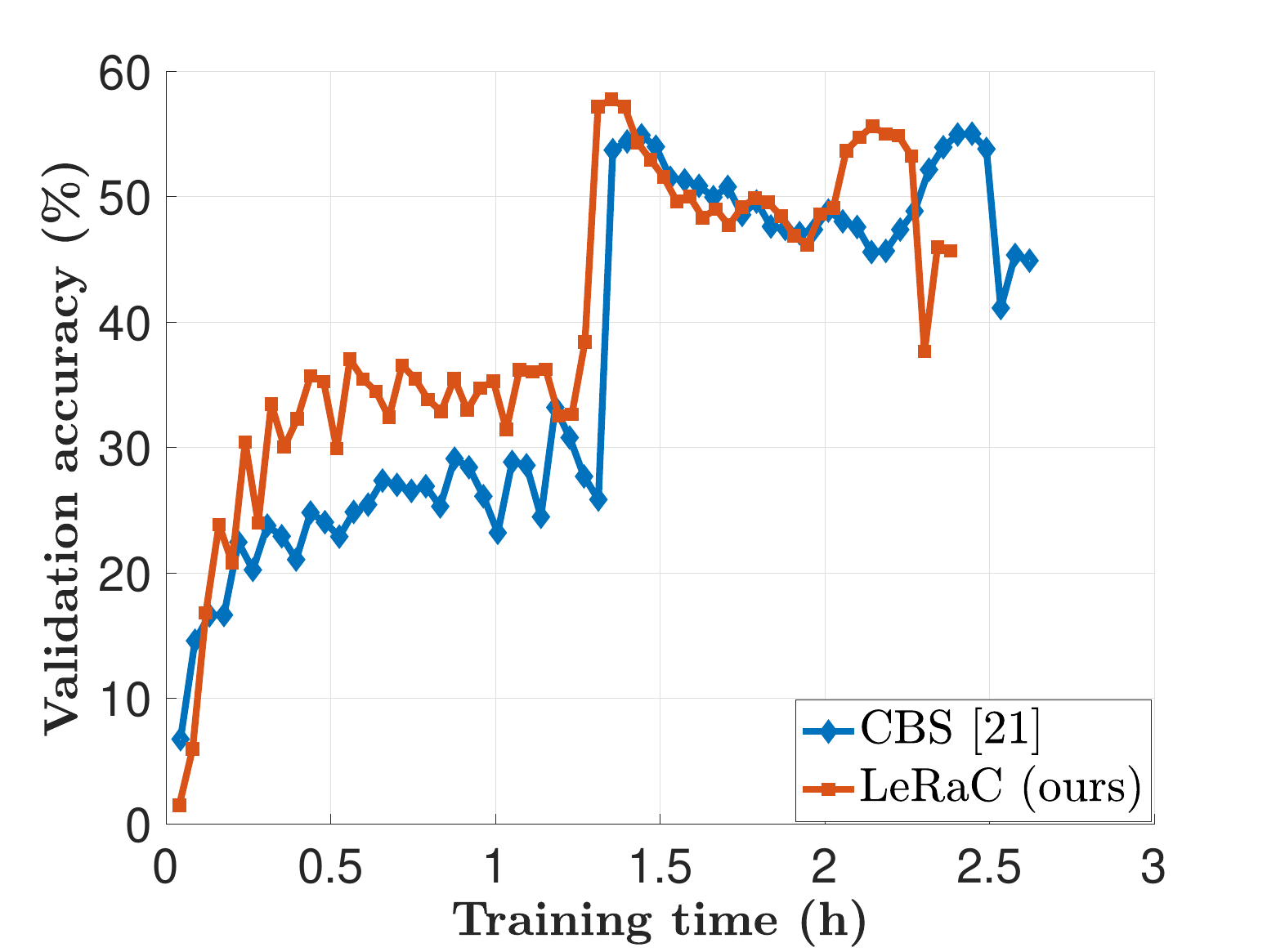}  
  \caption{ResNet-18 on Tiny ImageNet.}
  \label{fig:sub-third}
\end{subfigure}
\begin{subfigure}{.5\textwidth}
  \centering
  % include fourth image
  \includegraphics[width=.92\linewidth]{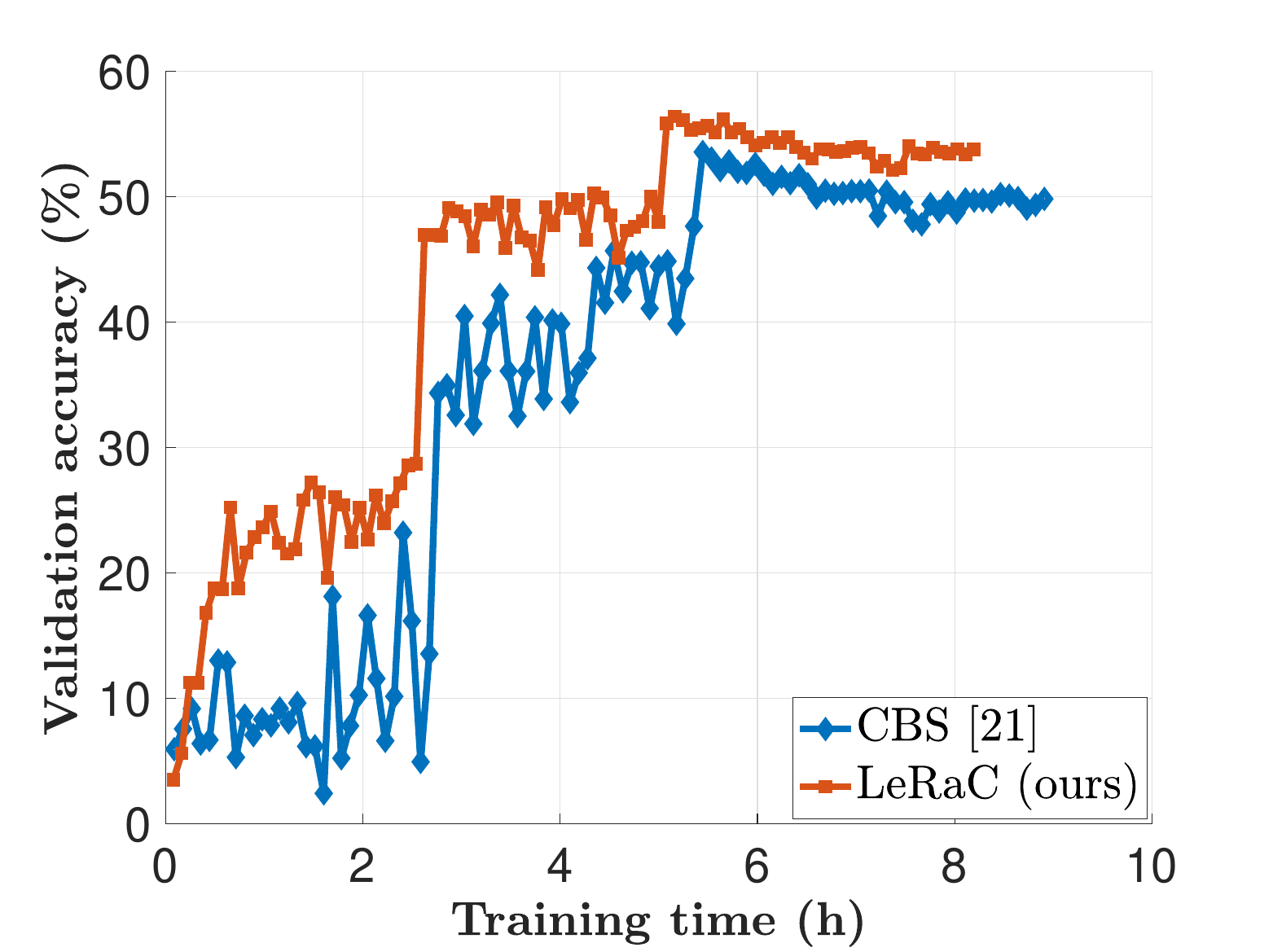}  
  \caption{Wide-ResNet-50 on Tiny ImageNet.}
  \label{fig:sub-fourth}
\end{subfigure}
\begin{subfigure}{.5\textwidth}
  \centering
  % include second image
  \includegraphics[width=.92\linewidth]{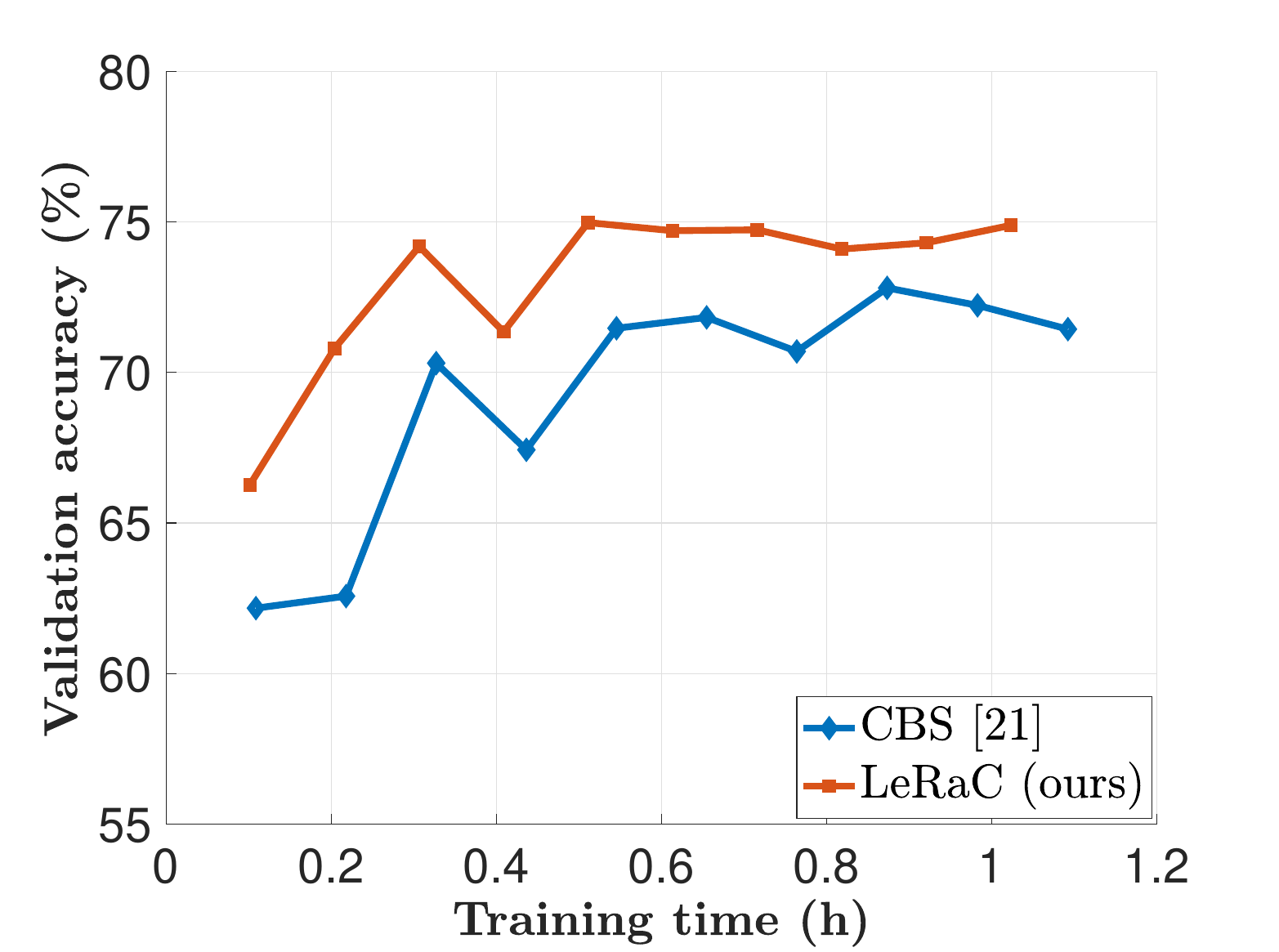}  
  \caption{BERT on BoolQ.}
  \label{fig:sub-second}
\end{subfigure}
\begin{subfigure}{.5\textwidth}
  \centering
  \includegraphics[width=.92\linewidth]{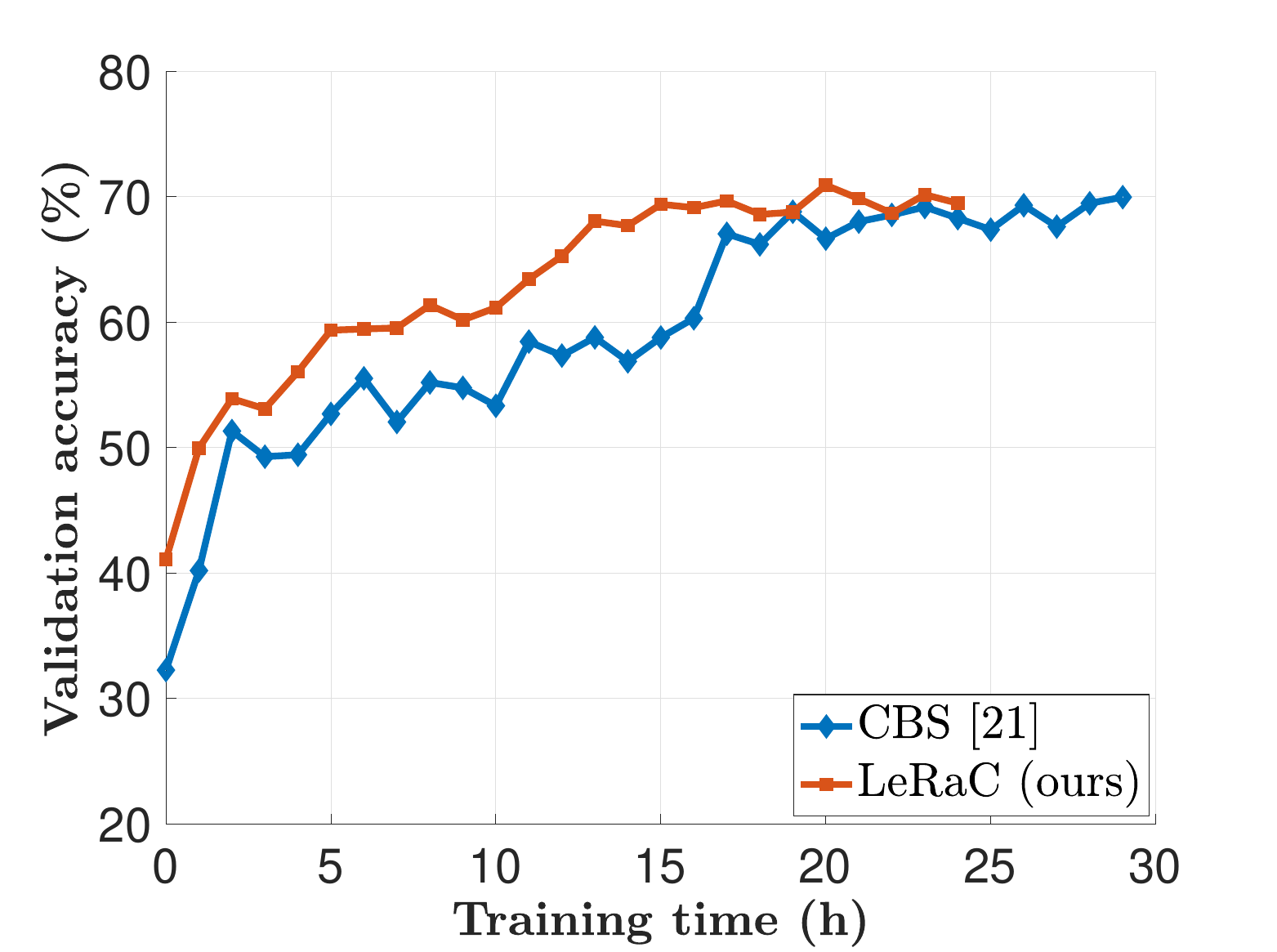}  
  \caption{SepTr on CREMA-D.}
  \label{fig:sub-first}
\end{subfigure}
\caption{Validation accuracy (on the y-axis) versus training time (on the x-axis) for four distinct architectures. The number of training epochs is the same for both LeRaC and CBS, the observable time difference being caused by the overhead of CBS due to the Gaussian kernel layers.\vspace{-0.1cm}}
\label{fig_train_time}
\end{figure*}

% \vspace{-0.1cm}
\noindent{\bf Speech classification.}
In Table~\ref{tab_text} (right side), we present the results obtained on the audio data sets, namely CREMA-D and ESC-50. %We employed a state-of-the-art method based on separable transformer blocks, SepTr \cite{Ristea-ARXIV-2022}, and a convolutional architecture, DenseNet-121 \cite{Gao-CVPR-2017}. 
We observe that the CBS strategy obtains lower results compared with the baseline in two cases (SepTr on CREMA-D and DenseNet-121 on ESC-50), while our method provides superior results for each and every case. By applying LeRaC on SepTr, we set a new state-of-the-art accuracy level ($70.95\%$) on the CREMA-D audio modality, surpassing the previous state-of-the-art value attained by \citet{Ristea-ARXIV-2022} with SepTr alone. When applied on DenseNet-121, LeRaC brings performance improvements higher than $1\%$, the highest improvement ($1.78\%$) over the baseline being attained on CREMA-D.

\noindent{\bf Significance testing.} 
To determine if the reported accuracy gains observed for LeRaC with respect to the baseline are significant, we apply McNemar / Cochran Q significance testing \citep{Dietterich-NC-1998} to the results reported in Table~\ref{tab_vision}, Table \ref{tab_multitask}, Table~\ref{tab_obj_detection} and Table~\ref{tab_text} on all 12 data sets. In 27 of 34 cases, we found that our results are significantly better than the corresponding baseline, at a p-value of $0.001$. This confirms that our gains are statistically significant in the majority of cases.

\begin{table*}[!t]
\caption{Average accuracy rates (in \%) over 5 runs for ResNet-18, Wide-ResNet-50 and CvT-13 (pre-trained) on CIFAR-10 and CIFAR-100 using different training regimes: conventional, CBS \citep{Sinha-NIPS-2020}, LSCL \citep{Dogan-ECCV-2020}, EfficientTrain \citep{Wang-ICCV-2023}, Self-taught \citep{Khan-arXiv-2024}, CLIP \citep{Khan-BigComp-2023}, LCDnet-CL \citep{Khan-RTIP-2023} and LeRaC (ours). The accuracy of the best training regime on each data set is highlighted in bold.}
  \label{tab_vision_more_baselines}
  \small{
  \begin{center}
  \begin{tabular}{|l|l|c|c|c|}
\hline
        Model  & Training Regime     & CIFAR-10 & CIFAR-100 \\
\hline
\hline
        \multirow{8}{*}{ResNet-18}              & conventional             & $89.20 \pm 0.43$ & $71.70 \pm 0.06$ \\
                                            & CBS \citep{Sinha-NIPS-2020} &  $89.53 \pm 0.22$ & $72.80 \pm 0.18$ \\
                                            & LSCL \citep{Dogan-ECCV-2020}  &  $88.28 \pm 0.14$ &  $68.42 \pm 0.25$ \\
                                            & EfficientTrain \citep{Wang-ICCV-2023}  &  $89.51 \pm 0.13$ & $\mathbf{72.83} \pm 0.12$ \\
                                            &  Self-taught~\citep{Khan-arXiv-2024} & $89.48 \pm 0.17$ & $72.10 \pm 0.32$ \\
                                            &  LCDnet-CL~\citep{Khan-RTIP-2023} & $89.36 \pm 0.38 $ & $71.06 \pm 0.27$\\
                                            &  CLIP~\citep{Khan-BigComp-2023} & $89.11 \pm 0.02$& $70.03 \pm 0.27$\\
                                               & LeRaC (ours)              & $\mathbf{89.56} \pm 0.16$ & $72.72 \pm 0.12$\\
\hline
         \multirow{8}{*}{Wide-ResNet-50}              & conventional             & $91.22 \pm 0.24$ & $68.14 \pm 0.16$ \\
                                                     & CBS \citep{Sinha-NIPS-2020} &  ${89.05} \pm 1.00$ &  ${65.73} \pm 0.36$ \\
                                                     & LSCL \citep{Dogan-ECCV-2020}  &  $88.28 \pm 0.14$ &  $72.59 \pm 0.25$ \\
                                            & EfficientTrain \citep{Wang-ICCV-2023} &  $91.03 \pm 0.28$ & $69.14 \pm 0.20$ \\
                                & Self-taught~\citep{Khan-arXiv-2024} & $91.00 \pm 0.24$ & $68.48 \pm 0.26$ \\
                                &  LCDnet-CL~\citep{Khan-RTIP-2023} & $91.38 \pm 0.18$ & $68.85 \pm 0.13$\\
                                &   CLIP~\citep{Khan-BigComp-2023} & $91.18 \pm 0.11$ & $68.13 \pm 0.39$\\
                                &                LeRaC (ours)             & $\mathbf{91.58} \pm 0.16$ & $\mathbf{69.38} \pm 0.26$ \\
                                \hline
         \multirow{8}{*}{CvT-13$_{\mbox{\scriptsize{pre-trained}}}$}              & conventional             &  $93.56 \pm 0.05$   &  $77.80 \pm 0.16$ \\
                                                     & CBS \citep{Sinha-NIPS-2020} &  $85.85 \pm 0.15$   &  $62.35 \pm 0.48$    \\
                                                     & LSCL \citep{Dogan-ECCV-2020}  &  $93.91 \pm 0.20$ &  $78.63 \pm 0.12$ \\
                                            & EfficientTrain \citep{Wang-ICCV-2023} &  $\mathbf{94.50} \pm 0.17$ & $78.20 \pm 0.34$ \\
                                &  Self-taught~\citep{Khan-arXiv-2024} & $92.25 \pm 0.22$ & $77.95 \pm 0.32$ \\
                                & LCDnet-CL~\citep{Khan-RTIP-2023} & $92.72 \pm 0.16 $ & $78.57 \pm 0.16$ \\
                                &  CLIP~\citep{Khan-BigComp-2023} & $92.61 \pm 0.36$ & $76.18 \pm 1.45$ \\
                                &                LeRaC (ours)             &  $94.15 \pm 0.03$   &  $\mathbf{78.93} \pm 0.05$  \\
\hline
      \end{tabular}
    \end{center}
    }
      % \vspace{-0.3cm}
\end{table*}

\noindent{\bf Training time comparison.}
For a particular model and data set, all training regimes are executed for the same number of epochs, for a fair comparison. However, the CBS strategy adds the smoothing operation at multiple levels inside the architecture, which increases the training time. To this end, we compare the training time (in hours) versus the validation error of CBS and LeRaC. For this experiment, we selected four neural models and illustrate the evolution of the validation accuracy over time in Figure~\ref{fig_train_time}. We underline that LeRaC introduces faster convergence times, being around 7-12\% faster than CBS. It is trivial to note that LeRaC requires the same time as the conventional regime. 

\subsection{More Comparative Results}

\noindent{\bf Comparing with domain-specific curriculum learning strategies.}
Although we consider CBS \citep{Sinha-NIPS-2020} as our closest competitor in terms of applicability across architectures and domains, there are domain-specific curriculum learning methods reporting promising results. To this end, we perform additional experiments on CIFAR-10 and CIFAR-100 with ResNet-18, Wide-ResNet-50 and CvT-13 (pre-trained), considering two recent curriculum learning strategies applied in the image domain, namely Label-Similarity Curriculum Learning (LSCL) \citep{Dogan-ECCV-2020} and EfficientTrain \citep{Wang-ICCV-2023}.

\begin{table*}[!t]
\caption{Average accuracy rates (in \%) over 5 runs on CIFAR-10, CIFAR-100 and Tiny ImageNet for CvT-13 based on different training regimes: conventional, LeRaC with logarithmic update, LeRaC with linear update,  and LeRaC with exponential update (proposed). The accuracy rates surpassing the baseline training regime are highlighted in bold.%\vspace{-0.3cm}
}
  \label{tab_lerac_schedule}
\small{
  \begin{center}
  \begin{tabular}{|l|l|c|c|c|}
    \hline
    Model       & Training Regime     & CIFAR-10  & CIFAR-100 & Tiny ImageNet \\
    \hline   
    \hline
    \multirow{4}{*}{CvT-13}             & conventional               &  $71.84 {\pm 0.37}$   &  $41.87 {\pm 0.16}$   & $33.38 {\pm 0.27}$ \\
    \cline{2-5}
                                  & LeRaC (logarithmic update) & $\mathbf{72.14} {\pm 0.13}$ & $\mathbf{43.37} {\pm 0.20}$ & $\mathbf{33.82} {\pm 0.15}$\\

                & LeRaC (linear update) & $\mathbf{72.49} {\pm 0.27}$ & $\mathbf{43.39} {\pm 0.14}$ & $\mathbf{33.86} {\pm 0.07}$\\
                 & LeRaC (exponential update)   &  $\mathbf{72.90} {\pm 0.28}$   &  $\mathbf{43.46} {\pm 0.18}$   & $\mathbf{33.95} {\pm 0.28}$ \\
    \hline
  \end{tabular}
  \end{center}
  }
  % \vspace{-0.35cm}
  
\end{table*}

\citet{Dogan-ECCV-2020} proposed LSCL, a strategy that relies on hierarchically clustering the classes (labels) based on inter-label similarities determined with the help of document embeddings representing the Wikipedia pages of the respective classes. The corresponding results shown in Table \ref{tab_vision_more_baselines} indicate that label-similarity curriculum is useful for CIFAR-100, but not for CIFAR-10. This suggests that the method needs a sufficiently large number of classes to benefit from the constructed hierarchy of classes. In contrast, LeRaC  does not rely on external components, such as the similarity measure used by \citet{Dogan-ECCV-2020} in their strategy. Another important limitation of LSCL \citep{Dogan-ECCV-2020} is its restricted use, \eg~LSCL is not applicable to regression tasks, where there are no classes. Therefore, we consider LeRaC as a more versatile alternative.

% \vspace{-0.1cm}
% \noindent{\bf Results on ImageNet-1K.} One question that naturally arises is if the results reported on ImageNet-200 in the main paper are transferable to the full ImageNet-1K. To this end, we conducted an experiment with ResNet-18 on ImageNet-1K to compare LeRaC with CBS \citep{Sinha-NIPS-2020} and EfficientTrain \citep{Wang-ICCV-2023}. Consistent with the results on ImageNet-200, the ImageNet-1K results reported in Table \ref{tab_ImageNet1K} show that LeRaC outperforms CBS \citep{Sinha-NIPS-2020}. Furthermore, LeRaC also surpasses a more recent approach reporting results in the same setting, namely EfficientTrain \citep{Wang-ICCV-2023}. 

%\noindent{\bf Comparing with other efficient training strategies.}

EfficientTrain is an alternative to CBS, which introduces a cropping operation in the Fourier spectrum of the inputs instead of blurring the activation maps. The method is not suitable for text data, so the comparison between EfficientTrain and LeRaC can only be performed in the image domain. Consequently, we compare with EfficientTrain \citep{Wang-ICCV-2023} on CIFAR-10 and CIFAR-100, and show the corresponding results in Table \ref{tab_vision_more_baselines}. Notably, our method surpasses EfficientTrain \citep{Wang-ICCV-2023} in 4 out of 6 evaluation scenarios. These results confirm the competitiveness of LeRaC in comparison to very recent methods, such as EfficientTrain \citep{Wang-ICCV-2023}. 

Aside from outperforming EfficientTrain and LSCL in the image domain, our method has another important advantage: it is generally applicable to any domain.

\noindent{\bf Comparing with data-level curriculum learning methods.} In Table~\ref{tab_vision_more_baselines}, we also compare LeRaC with three data-level curriculum learning methods \citep{Khan-arXiv-2024, Khan-BigComp-2023, Khan-RTIP-2023}. These methods share a common framework, where a scoring function ranks samples based on their difficulty, and a pacing function determines the timing for introducing new batches during training. \citet{Khan-arXiv-2024} examine various pacing functions and classify scoring functions into two categories: self-taught and transfer-scoring functions. Self-taught functions involve training a model on a subset of data batches and then using this model to assess the difficulty of examples. In contrast, transfer-scoring functions utilize a pre-trained model for this purpose. For the results reported in Table~\ref{tab_vision_more_baselines} for \citet{Khan-arXiv-2024}, we use the self-taught scoring function and a linear pacing function. To compare with \citet{Khan-RTIP-2023}, we use a transfer-scoring function and a ResNet-50 model pre-trained on ImageNet. For \citet{Khan-BigComp-2023}, aside from using the pre-trained model for assessing the difficulty of the samples, we also remove the least significant samples during training.

The results reported in Table~\ref{tab_vision_more_baselines} indicate that LeRaC outperforms the data-level curriculum learning methods. We note that these methods were exclusively tested on crowd density estimation tasks, which could explain why their effectiveness might not generalize to different types of tasks. For instance, the method described by \citet{Khan-BigComp-2023} is suboptimal even when compared with conventional training, suggesting that the strategy of removing easy examples is not always effective for image classification tasks.

\begin{figure}[!t]
\begin{center}
\centerline{\includegraphics[width=1.0\linewidth]{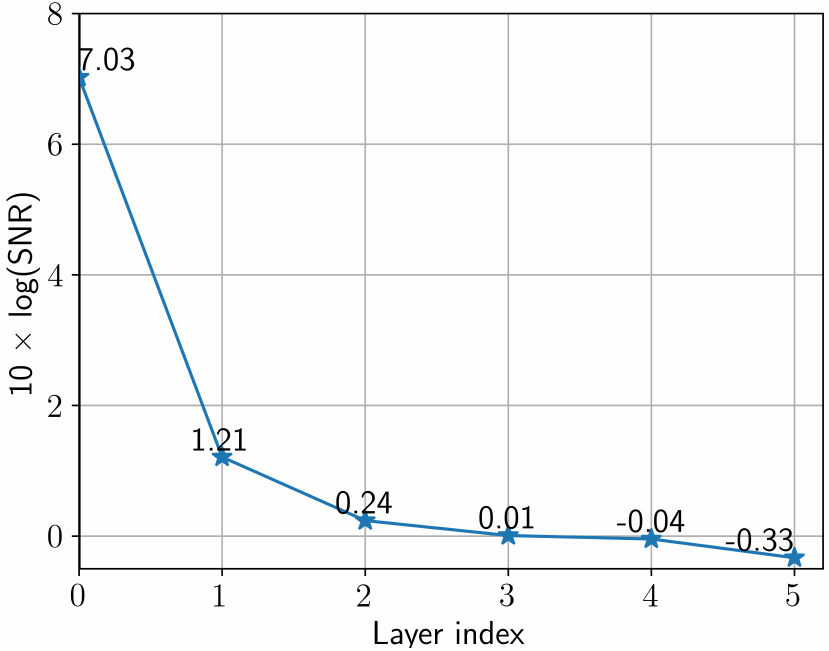}}
\caption{Average SNR of the feature maps at each layer of the randomly initialized LeNet architecture. The SNR at each layer is averaged for 100 randomly picked images from the CIFAR-100 data set. For the later layers, the SNR is negative because the signal is dominated by noise.}
\label{fig_snr_lenet}
\end{center}
\end{figure} 

\subsection{Ablation Studies}

\noindent{\bf Comparing different schedulers.} 
We first aim to establish if the exponential learning rate scheduler proposed in Eq.~(\ref{eq_update_exp}) is a good choice. To test this out, we select the CvT-13 model and change the LeRaC regime to use linear or logarithmic updates of the learning rates. The corresponding results are shown in Table~\ref{tab_lerac_schedule}. We observe that both alternative schedulers obtain performance gains, but our exponential learning rate scheduler brings higher gains on all three data sets. We thus conclude that the update rule defined in Eq.~(\ref{eq_update_exp}) is a sound option.

%\noindent{\bf Motivation for exponential scheduler.} 

Our previous ablation study %from the main paper
shows that the exponential scheduler leads to higher gains than the linear or the logarithmic schedulers. In general, a suitable scheduler is one that adjusts the learning rate at each layer proportionally to the estimated signal-to-noise drop from one layer to the next. To understand how the average SNR drops from one neural layer to the next, we plot the average SNR of the features maps at each layer of the randomly initialized LeNet architecture, computed over 100 images from CIFAR-100, in Figure~\ref{fig_snr_lenet}. As anticipated, the average SNR decreases along with the layer index. Notably, we observe that the drop in SNR follows an exponential trend. This can explain why the exponential scheduler is a more suitable choice. 

\begin{figure}[!t]
\begin{center}
\centerline{\includegraphics[width=1.0\linewidth]{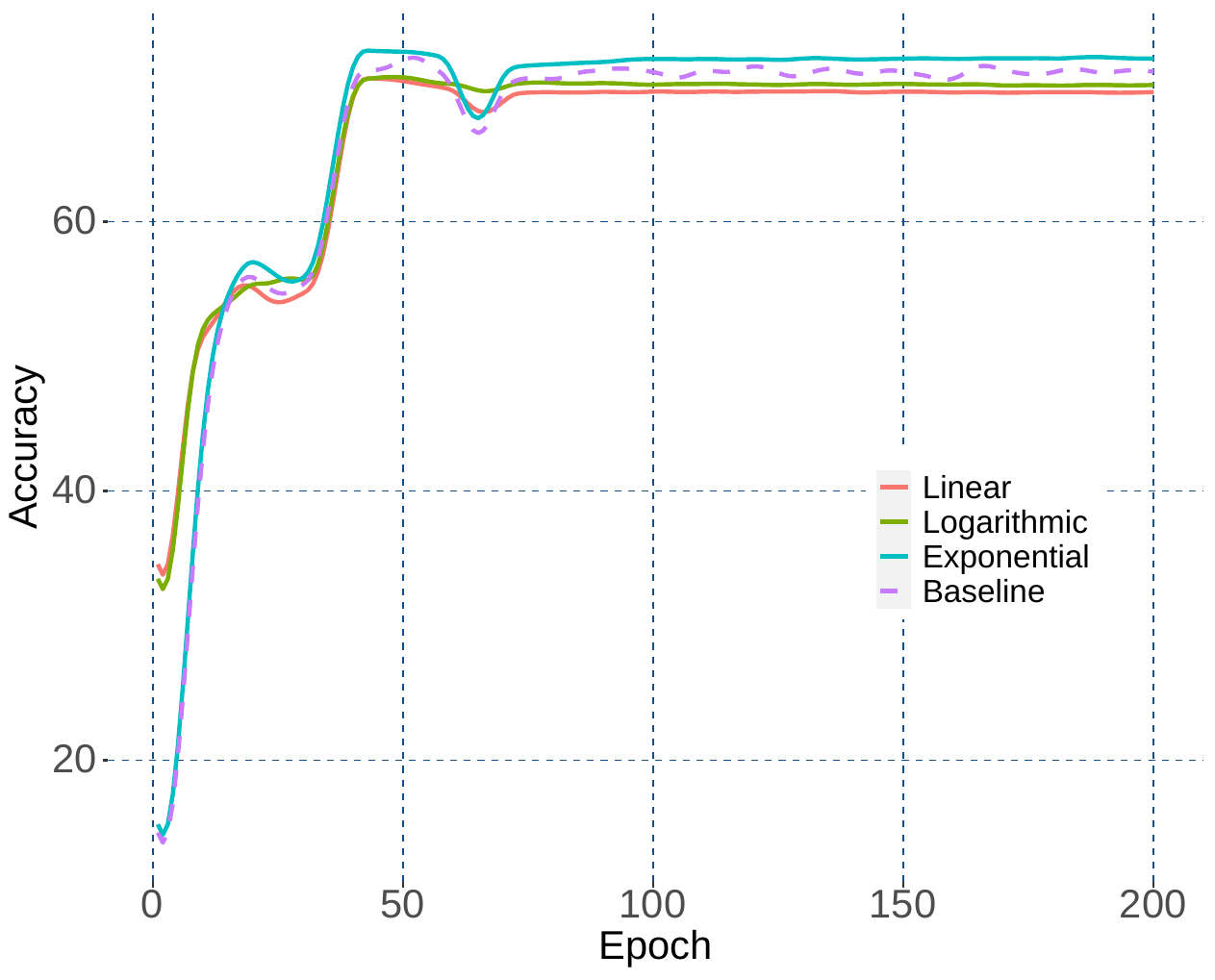}}
\caption{Test accuracy (on the y-axis) versus training time (on the x-axis) for ResNet-18 on CIFAR-100 with various curriculum schedulers. The dashed line corresponds to the conventional regime, while the continuous lines correspond to LeRaC with various schedulers. Best viewed in color.}
\label{fig_r18_cif100_schedulers}
\end{center}
\end{figure}

\begin{figure}[!t]
\begin{center}
\centerline{\includegraphics[width=1.0\linewidth]{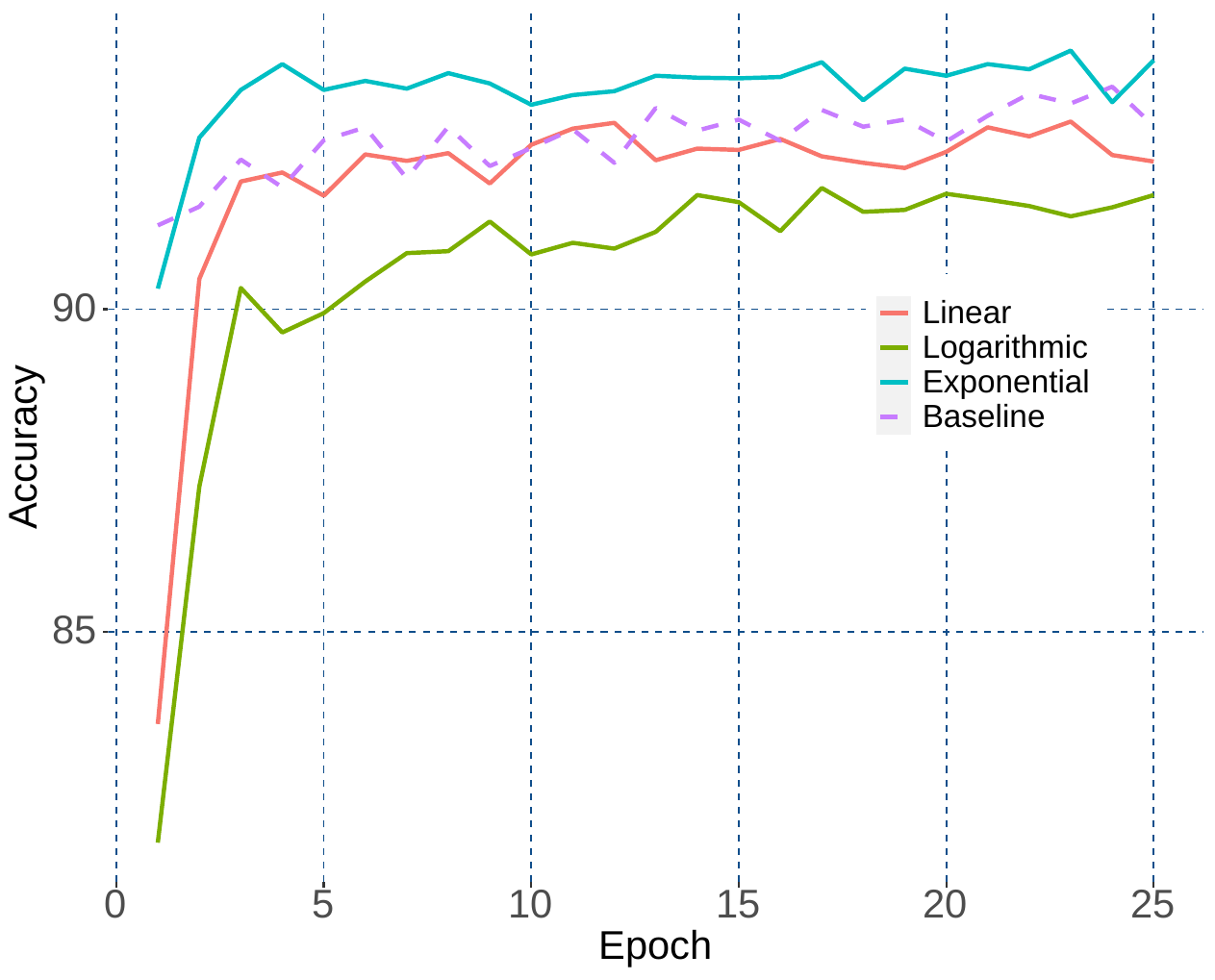}}
\caption{Test accuracy (on the y-axis) versus training time (on the x-axis) for the pre-trained CvT-13 on CIFAR-10 with various curriculum schedulers. The dashed line corresponds to the conventional regime, while the continuous lines correspond to LeRaC with various schedulers. Best viewed in color.}
\label{fig_cvt_cif10_schedulers}
\end{center}
\end{figure}

To further justify our preference towards the exponential scheduler, we analyze the training progress of the ResNet-18 and the pre-trained CvT-13 models using various schedulers (logarithmic, linear, exponential) for LeRaC. Figure \ref{fig_r18_cif100_schedulers} shows the results for ResNet-18, while Figure \ref{fig_cvt_cif10_schedulers} illustrates the results for CvT-13. In both cases, the exponential scheduler leads to a better training progress than the conventional regime, but the linear and logarithmic schedulers are not as good. These results further confirm that the exponential scheduler is optimal.

\begin{table*}[t]
 \caption{Average accuracy rates (in \%) over 5 runs for ResNet-18 and Wide-ResNet-50 on CIFAR-100 based on different ranges for the initial learning rates. The accuracy rates surpassing the baseline training regime are highlighted in bold.}
  \label{tab_various_ranges}
  \small{
  \renewcommand{\arraystretch}{1.1}
  \begin{center}
  \begin{tabular}{|l|c|c|c|}
\hline
Training  Regime & {$\eta_1^{(0)}$-$\eta_n^{(0)}$}     & {ResNet-18} & {Wide-ResNet-50} \\
\hline
\hline
conventional    & $10^{-1}$-$10^{-1}$ & $\;\;\;\;71.70 {\pm0.06}\;\;\;\;$ & $68.14 {\pm0.16}$ \\
\hline
\multirow{7}{*}{LeRaC (ours)} & $10^{-1}$-$10^{-6}$             & $\mathbf{72.48} {\pm0.10}$ & $\mathbf{68.64} {\pm0.52}$\\
  & $10^{-1}$-$10^{-7}$             & $\mathbf{72.52} {\pm0.17}$ & $\mathbf{69.25} {\pm0.37}$\\
  & $10^{-1}$-$10^{-8}$             & $\mathbf{72.72} {\pm0.12}$ & $\mathbf{69.38} {\pm0.26}$\\
 & $10^{-1}$-$10^{-9}$             & $\mathbf{72.29} {\pm0.38}$ & $\mathbf{69.26} {\pm0.27}$\\
 & $10^{-1}$-$10^{-10}$             & $\mathbf{72.45} {\pm0.25}$ & $\mathbf{69.66} {\pm0.34}$\\
  & $10^{-2}$-$10^{-8}$             & $\mathbf{72.41} {\pm0.08}$ & $\mathbf{68.51} {\pm0.52}$\\
  & $10^{-3}$-$10^{-8}$             & $\mathbf{72.08} {\pm0.19}$ & $\mathbf{68.71} {\pm0.47}$\\
    \hline
    \end{tabular}
\end{center}
    }
      %\vspace{-0.3cm}
   
\end{table*}

% \vspace{-0.1cm}
\noindent{\bf Varying value ranges for initial learning rates.}
All our hyperparameters are either fixed without tuning or tuned on the validation data. In this ablation experiment, we present results with LeRaC using multiple ranges for $\eta_1^{(0)}$ and $\eta_n^{(0)}$ to demonstrate that LeRaC is sufficiently stable with respect to suboptimal hyperparameter choices. We carry out experiments with ResNet-18 and Wide-ResNet-50 on CIFAR-100. We report the corresponding results in Table~\ref{tab_various_ranges}. We observe that all hyperparameter configurations lead to surpassing the baseline regime. This indicates that LeRaC can bring performance gains even outside the optimal learning rate bounds, demonstrating low sensitivity to suboptimal hyperparameter tuning.

\begin{table*}[t]
    \caption{Average accuracy rates (in \%) over 5 runs for ResNet-18 and Wide-ResNet-50 on CIFAR-100 using the LeRaC regime until iteration $k$, while varying $k$. The accuracy rates surpassing the baseline training regime are highlighted in bold.%\vspace{-0.3cm}
    }
  \label{tab_var_k}
  \small{
  \begin{center}
  \begin{tabular}{|l|c|c|c|}
\hline
Training Regime   & $k$  & ResNet-18 & Wide-ResNet-50\\
\hline
\hline
 conventional           & - & $\;\;\;\;71.70 {\pm0.06}\;\;\;\;$ & $68.14 {\pm0.16}$\\
     \hline
                                                          & 5  & $\mathbf{73.04} {\pm0.09}$ & $\mathbf{68.86} {\pm0.76}$\\
                                                           & 6  & $\mathbf{72.87} {\pm0.07}$ & $\mathbf{69.78} {\pm0.16}$\\
                                                LeRaC (ours)           & 7  & $\mathbf{72.72} {\pm0.12}$ & $\mathbf{69.38} {\pm0.26}$\\
                                                          & 8  & $\mathbf{73.50} {\pm0.16}$ & $\mathbf{69.30} {\pm0.18}$\\
                                                         & 9  & $\mathbf{73.29} {\pm0.28}$ & $\mathbf{68.94} {\pm0.30}$\\
\hline
      \end{tabular}
    \end{center}
    }
      % \vspace{-0.3cm}

\end{table*}

\begin{table*}[t]
\caption{Average accuracy rates (in \%) over 5 runs for ResNet-18 and Wide-ResNet-50 on CIFAR-100, as well as SepTr on CREMA-D, based on different training regimes: conventional, anti-LeRaC and LeRaC. The accuracy of the best training regime in each experiment is highlighted in bold.}
  \label{tab_anti}
  \small{
  \setlength\tabcolsep{2.9pt}
  \begin{center}
  \begin{tabular}{|l|l|l|c|}
\hline
        Data Set & Model  & Training Regime     & Accuracy \\
\hline
\hline
        \multirow{6}{*}{\vspace{-0.15cm}CIFAR-100}  &               & conventional             & $71.70\!\pm\!0.06$ \\
                                & ResNet-18     & anti-LeRaC             & $71.24\!\pm\!0.11$\\
                                &               & LeRaC (ours)              & $\mathbf{72.72}\!\pm\!0.12$\\
    \cline{2-4}
    &               & conventional             & $68.14\!\pm\!0.16$ \\
                       & Wide-ResNet-50     & anti-LeRaC             & $67.47\!\pm\!0.15$\\
                                &               & LeRaC (ours)              & $\mathbf{69.38}\!\pm\!0.26$\\
\hline
        &      & conventional                   &   $70.47\!\pm\!0.67$     \\
    CREMA-D & SepTr     & anti-LeRaC                &   $68.33\!\pm\!0.61$    \\
    &     & LeRaC (ours)                    & $\mathbf{70.95}\!\pm\!0.56$   \\
\hline
      \end{tabular}
    \end{center}
    }
      % \vspace{-0.3cm}
    
\end{table*}

% \vspace{-0.1cm}
\noindent{\bf Varying $\mathbf{k}$.}
In Table \ref{tab_var_k}, we present additional results with ResNet-18 and Wide-ResNet-50 on CIFAR-100, considering various values for $k$ (the last iteration for our training regime). We observe that all configurations surpass the baselines on CIFAR-100. Moreover, we observe that the optimal values for $k$ ($k=7$ for ResNet-18 and $k=7$ for Wide-ResNet-50) obtained on the validation set are not the values producing the best results on the test set. This confirms that we did not overfit the hyperparameters of LeRaC.

\noindent{\bf Anti-curriculum.}
Since our goal is to perform curriculum learning (from easy to hard), we restrict the settings for $\eta_j$, $\forall j \in \{1,2,...,n \}$, such that deeper layers start with lower learning rates. However, another strategy is to consider the opposite setting, where we use higher learning rates for deeper layers. If we train later layers at a faster pace (anti-curriculum), we conjecture that the later layers get adapted to the noise from the early layers, which could likely lead to local optima or difficult training (due to the need of readapting to the earlier layers, once these layers start learning useful features).
We tested this approach (anti-LeRaC), which belongs to the category of anti-curriculum learning strategies \citep{Soviany-IJCV-2022}, in a set of new experiments with ResNet-18 and Wide-ResNet-50 on CIFAR-100, as well as SepTr on CREMA-D. We report the corresponding results with LeRaC and anti-LeRaC in Table~\ref{tab_anti}. Although anti-curriculum, \eg~hard negative sample mining, was shown to be useful in other tasks \citep{Soviany-IJCV-2022}, our results indicate that learning rate anti-curriculum attains inferior performance compared with our approach. Furthermore, anti-LeRaC is also below the conventional regime, confirming our conjecture regarding this strategy.

% \vspace{-0.1cm}
\noindent{\bf Summary.} Notably, our ablation results show that the majority of hyperparameter configurations tested for LeRaC lead to outperforming the conventional regime, demonstrating the stability of LeRaC. We present additional %ablation
experiments in Appendix~\ref{app_additional_exp}.

\section{Discussion}

\noindent{\bf Interaction with optimization algorithms.}
Throughout our experiments, we always keep using the same optimizer for a certain neural model, for all training regimes (conventional, CBS, LeRaC). The best optimizer for each neural model is established for the conventional training regime. We underline that our initial learning rates and scheduler are used independently of the optimizers. Although our learning rate scheduler updates the learning rates at the beginning of every iteration, we did not observe any stability or interaction issues with any of the optimizers (SGD, Adam, AdaMax, AdamW).

\noindent{\bf Interaction with other curriculum learning strategies.}
Our simple and generic curriculum learning scheme can be integrated into any model for any task, not relying on domain or task dependent information, \eg~the data samples. In Table \ref{tab_extra} from Appendix \ref{app_additional_exp}, we show that combining LeRaC and CBS can boost performance. In a similar fashion, LeRaC can be combined with data-level curriculum strategies for improved performance. We leave this exploration for future work.

\noindent{\bf Interaction with other learning rate schedulers.}
Whenever a learning rate scheduler is used for training a model in our experiments, we simply replace the scheduler with LeRaC until epoch $k$. For example, all the baseline CvT results are based on linear warm-up with cosine annealing, this being the recommended scheduler for CvT \citep{Wu-ICCV-2021}. When we introduce LeRaC, we simply deactivate alternative schedulers between epochs $0$ and $k$. In general, we recommend deactivating other schedulers while using LeRaC for simplicity in avoiding stability issues.

\noindent{\bf Limitations of our work.}
One limitation is the need to disable other learning rate schedulers while using LeRaC. We already tested this scenario with linear warm-up with cosine annealing, which is removed when using LeRaC, observing consistent performance gains (see Table \ref{tab_vision}). However, disabling alternative learning rate schedulers might bring performance drops in other cases. Hence, this has to be decided on a case by case basis. Another limitation is the possibility of encountering longer training times or poor convergence when the hyperparameters are not properly configured. We recommend hyperparameter tuning on the validation set to avoid this outcome. % An additional limitation is that we tested our approach on mainstream classification tasks involving mainstream classification losses (multi-class or binary cross-entropy). We leave the integration with additional losses for future work.

\section{Conclusion}

In this paper, we introduced a novel model-level curriculum learning approach that is based on starting the training process with increasingly lower learning rates per layer, as the layers get closer to the output. We conducted comprehensive experiments on 12 data sets from three domains (image, text and audio), considering multiple neural architectures (CNNs, RNNs and transformers), to compare our novel training regime (LeRaC) with a state-of-the-art regime (CBS~\citep{Sinha-NIPS-2020}), as well as the conventional training regime (based on early stopping and reduce on plateau). The empirical results demonstrate that LeRaC is significantly more consistent than CBS, perhaps being one of the most versatile curriculum learning strategy to date, due to its compatibility with multiple neural models and its usefulness across different domains. Remarkably, all these benefits come for free, \ie~LeRaC does not add any extra time over the conventional approach.

% \noindent{\bf Acknowledgements.} Work funded by UEFISCDI through project number PN-III-P1-1.1-TE-2019-0235.

\section*{Declarations}

\textbf{Funding.} This work was supported by a grant of the Romanian Ministry of Education and Research, CNCS - UEFISCDI, project number PN-III-P2-2.1-PED-2021-0195, within PNCDI III.

\vspace{0.2cm}
\noindent
\textbf{Conflict of interest.} The authors have no conflicts of interest to declare that are relevant to the content of this article.

\vspace{0.2cm}
\noindent
\textbf{Availability of data and materials.} The data sets are publicly available online.

\vspace{0.2cm}
\noindent
\textbf{Code availability.} The code has been made publicly available for non-commercial use at \url{https://github.com/CroitoruAlin/LeRaC}.

\appendix

%\begin{appendices}

\makeatletter
\@addtoreset{theorem}{section}
\makeatother

\section{Theoretical Proof}
\label{app_proof}
The motivation behind using LeRaC stems from our conjecture stating that the level of noise inside features gradually increases with every layer of a neural network. Regardless of the type of layer (convolutional, transformer or fully connected), the operation performed inside a neural layer boils down to matrix or vector multiplications. To this end, we set out to demonstrate that the signal resulting from the multiplication of two signals has a lower signal-to-noise ratio (SNR) than the multiplied factors. We start with the definition of the variance of a signal, which is given below:
\begin{definition}\label{def_var}
The variance of a signal $s$ is given by:
\begin{equation}\label{eq_var}
\Var(s)=E[s^2] - E[s]^2.
\end{equation}
\end{definition}

From Definition \ref{def_var}, it results that the expected value of $s^2$, which represents the power of signal $s$, is equal to:
\begin{equation}\label{eq_power}
E[s^2] = E[s]^2 + \Var(s) = \mu^2_s + \sigma^2_s,
\end{equation}
where $\mu_s$ is the mean of $s$, and $\sigma^2_s$ is the variance of $s$. We use Eq.~(\ref{eq_power}) to define the SNR of a signal as follows:
\begin{definition}\label{def_snr}
The signal-to-noise ratio (SNR) of a signal $s=u+z$, where $u$ is the clean signal and $z$ is the noise component, is the ratio between the power of $u$ and the power of $z$, which is given by:
\begin{equation}\label{eq_snr}
\SNR(s)=\frac{E[u^2]}{E[z^2]}=\frac{\mu^2_u + \sigma^2_u}{\mu^2_z + \sigma^2_z},
\end{equation}
where $\mu_u$ and $\mu_z$ are the means of $u$ and $z$, and $\sigma^2_u$ and $\sigma^2_z$ are the variances of $u$ and $z$, respectively.
\end{definition}

% Examples given as input to neural networks are often standardized by subtracting the mean and dividing by the standard deviation. Moreover, 
% 

The noise contained by data samples given as input to neural networks is usually uncorrelated, \eg~the noise in images is assumed to come from a random normal distribution of zero mean. Moreover, the weights of a neural network are usually initialized by sampling them from a random normal distribution of zero mean \citep {Glorot-AISTATS-2010}. Hence, without loss of generality, we can naturally assume that the noise component has zero mean. This means that we can simplify Eq.~(\ref{eq_snr}) to:
\begin{equation}\label{eq_snr_simple}
\SNR(s)=\frac{\mu^2_u + \sigma^2_u}{\sigma^2_z}.
\end{equation}

% An example given as input to a neural network and the initial weights of the respective neural network are not correlated under any practical circumstances. 

If the power of the signal $u$ is higher than the power of the noise $z$, then $\SNR(s)$ is higher than $1$. If the signal is dominated by noise, then $\SNR(s)$ is between $0$ and $1$. Note that the SNR does not take negative values. To avoid discussing edge cases, we assume that the SNR of any signal is always defined, \ie~the power of the noise is never $0$. 

% We next provide our demonstration for this practical case.

\begin{theorem}\label{prop_snr_apx}
Let $s_1=u_1+z_1$ and $s_2=u_2+z_2$ be two signals, where $u_1$ and $u_2$ are the clean components, and $z_1$ and $z_2$ are the noise components. The signal-to-noise ratio of the product between the two signals is lower than the signal-to-noise ratios of the two signals, \ie:
\begin{equation}
\SNR(s_1\cdot s_2) \leq \SNR(s_i), \forall i \in \{1, 2\}.
\end{equation}
\end{theorem}

\begin{proof} To demonstrate our theorem, we rely on the formula of variance for the sum of two signals with zero mean: 
\begin{equation}\label{eq_var_sum}
\Var(s_1 + s_2) = \Var(s_1) + \Var(s_2).
\end{equation}
We also rely on the formula of variance for the product of two signals: 
\begin{equation}\label{eq_var_prod}
\begin{split}
\Var(s_1\cdot s_2) = &\Var(s_1) \cdot \Var(s_2) + \Var(s_1) \cdot E[s_2]^2 \\
&+  \Var(s_2) \cdot E[s_1]^2.
\end{split}
\end{equation}

Let $s$ denote the product of the two signals, \ie~$s=s_1 \cdot s_2$.
Expanding the signals $s_1$ and $s_2$ leads to the following formulation of $s$:
\begin{equation}
\begin{split}
s &= s_1 \cdot s_2 = (u_1 + z_1) \cdot (u_2 + z_2) \\
&= u_1 \cdot u_2 + u_1 \cdot z_2 + u_2 \cdot z_1 + z_1 \cdot z_2,
\end{split}
\end{equation}
where the clean component is $u=u_1 \cdot u_2$, and the noise component is $z=u_1 \cdot z_2 + u_2 \cdot z_1 + z_1 \cdot z_2$. Hence, $s=u+z$.

An example given as input to a neural network and the initial weights of the respective neural network are not correlated under any practical circumstances. Hence, without loss of generality, we can assume that the signals $s_1$ and $s_2$ are independent of each other, \ie~their covariance is equal to $0$. This assumption allows us to simplify the signal power of $u$ to:
\begin{equation}\label{eq_pow_clean}
\begin{split}
E[u^2] &= E[u_1^2 \cdot u_2^2] = E[u_1^2] \cdot E[u_2^2] \\
&= \left(\mu^2_{u_1} + \sigma^2_{u_1} \right) \cdot \left(\mu^2_{u_2} + \sigma^2_{u_2}\right).    
\end{split}
\end{equation}

The signal power of $z$ is given by:
\begin{equation}\label{eq_pow_noise1}
\begin{split}
E[z^2] = E[z]^2 + \Var(z) = \Var(z),
\end{split}
\end{equation}
since the noise is of zero mean, \ie~$E[z] = 0$. By employing Eq.~(\ref{eq_var_sum}), we can compute the power of $z$ as follows:
\begin{equation}\label{eq_pow_noise2}
\begin{split}
\!\!E[z^2] &= \Var(z) = \Var(u_1 \cdot z_2 + u_2 \cdot z_1 + z_1 \cdot z_2)\\
&= \Var(u_1 \cdot z_2) + \Var(u_2 \cdot z_1) + \Var(z_1 \cdot z_2).
\end{split}
\end{equation}
By applying Eq.~(\ref{eq_var_prod}) in Eq.~(\ref{eq_pow_noise2}), and considering that $z_1$ and $z_2$ have zero mean, we obtain:
\begin{equation}\label{eq_pow_noise_components}
\begin{split}
\Var(u_1 \cdot z_2) &= \left(\mu^2_{u_1} + \sigma^2_{u_1} \right) \cdot \sigma^2_{z_2},\\
\Var(u_2 \cdot z_1) &= \left(\mu^2_{u_2} + \sigma^2_{u_2} \right) \cdot \sigma^2_{z_1},\\
\Var(z_1 \cdot z_2) &= \sigma^2_{z_1} \cdot \sigma^2_{z_2}.\\
\end{split}
\end{equation}

Replacing Eq.~(\ref{eq_pow_clean}) and Eq.~(\ref{eq_pow_noise_components}) inside Definition~\ref{def_snr} leads to the following expression of the signal-to-noise ratio of signal $s$:
\begin{equation}\label{eq_snr_prod}
\begin{split}
\SNR&(s)=\frac{E[u^2]}{E[z^2]} \\
&=\!\frac{\left(\mu^2_{u_1} + \sigma^2_{u_1} \right) \cdot \left(\mu^2_{u_2} + \sigma^2_{u_2}\right)}{\left(\mu^2_{u_1}\!+\!\sigma^2_{u_1} \right)\!\cdot\! \sigma^2_{z_2}\!+\!\left(\mu^2_{u_2}\!+\!\sigma^2_{u_2} \right)\!\cdot\!\sigma^2_{z_1}\!+\!\sigma^2_{z_1}\!\cdot\!\sigma^2_{z_2}}\\
&=\frac{\left(\mu^2_{u_1} + \sigma^2_{u_1} \right) \cdot \left(\mu^2_{u_2} + \sigma^2_{u_2}\right)}
{\sigma^2_{z_1}\!\cdot\!\sigma^2_{z_2}\!\cdot\! \left(
\frac{\mu^2_{u_1}\!+\!\sigma^2_{u_1}}{\sigma^2_{z_1}}
\!+\!
\frac{\mu^2_{u_2}\!+\!\sigma^2_{u_2}}{\sigma^2_{z_2}}
\!+\!1
\right)} \\
&=\frac{\frac{\mu^2_{u_1} + \sigma^2_{u_1} }{\sigma^2_{z_1}} \cdot \frac{\mu^2_{u_2} + \sigma^2_{u_2}}{\sigma^2_{z_2}}}
{\frac{\mu^2_{u_1}\!+\!\sigma^2_{u_1}}{\sigma^2_{z_1}}
\!+\!
\frac{\mu^2_{u_2}\!+\!\sigma^2_{u_2}}{\sigma^2_{z_2}}
\!+\!1}\\
&=\frac{\SNR(s_1) \cdot \SNR(s_2)}{\SNR(s_1) + \SNR(s_2) + 1}.
\end{split}
\end{equation}
To simplify our notations in the remainder of this proof, we define $a=\SNR(s_1)$ and $b=\SNR(s_2)$. By introducing these notations in Eq.~(\ref{eq_snr_prod}), we obtain the following:
% and applying Definition~\ref{def_snr} to $s_1$ and $s_2$, we can express the powers of the clean signals $u_1$ and $u_2$ as follows:
% \begin{equation}\label{eq_pow_clean_notation}
% \begin{split}
% \sigma^2_{u_1} &= a \cdot \sigma^2_{z_1},\;\; \sigma^2_{u_2} = b \cdot \sigma^2_{z_2}.
% \end{split}
% \end{equation}
% Next, we replace the powers of the clean signals defined in Eq.~\eqref{eq_pow_clean_notation} 
% into Eq.~\eqref{eq_snr_prod}, obtaining the following:
\begin{equation}\label{eq_snr_relation}
\begin{split}
\SNR(s)= \frac{a \cdot b}{a+b+1}.
\end{split}
\end{equation}
Now, it remains to prove that:
\begin{equation}\label{eq_ineq1}
\frac{a \cdot b}{a+b+1} \leq a,\;\;
\frac{a \cdot b}{a+b+1} \leq b.
\end{equation}
However, since $a$ and $b$ are commutable in Eq.~(\ref{eq_snr_relation}), it is sufficient to prove only one of the inequalities. We choose to provide the complete proof for the first inequality in Eq.~(\ref{eq_ineq1}) (as the proof for the other is analogous). We consider two separate cases, $a=0$ and $a>0$.

\textbullet\ {Case $(i)$:} When $a=0$, we obtain the following inequality:
\begin{equation}\label{eq_ineq2}
\frac{0}{b+1} \leq 0,
\end{equation}
which clearly holds for any $b \geq 0$.

\begin{table*}[t]
\caption{Distances between feature maps at epoch $k=0$ and feature maps after the final epoch for ResNet-18 on CIFAR-10, while using the conventional training regime. Distances are independently computed for the first and last convolutional layers.}
  \label{tab_noise_quant}
  \small{
  \begin{center}
  \begin{tabular}{|l|c|c|}
    \hline
    \multirow{2}{*}{Training Regime}     & \multicolumn{2}{c|}{Distance} \\
    \cline{2-3}
    & First Conv Layer & Last Conv Layer \\
    \hline
    \hline
     conventional             & $38.36$ & $709.93$\\
    \hline
  \end{tabular}
    \end{center}
    }
      % \vspace{-0.3cm}
\end{table*}

\textbullet\ {Case $(ii)$:} When $a>0$, we can divide both terms of the inequality by $a$ and arrive to: 
\begin{equation}\label{eq_ineq3}
\frac{b}{a+b+1} \leq 1.
\end{equation}
Next, we multiply both terms by $a+b+1$, obtaining that:
\begin{equation}\label{eq_ineq4}
b \leq a+b+1.
\end{equation}
We can subtract $b$ from both terms and obtain the following:
\begin{equation}\label{eq_ineq5}
0 \leq a+1.
\end{equation}
Since $a>0$, it results that Eq.~(\ref{eq_ineq5}) is true. Moreover, the inequality is strict when $a>0$.
This concludes our proof.
\end{proof}

\begin{corollary}\label{prop_snr_multi}
Let $\{s_1, s_2,...s_n\}$ be a set of $n$ signals, where each signal $s_i=u_i+z_i$ is formed of a clean component $u_i$ and a noise component $z_i$. The following equation is true:
\begin{equation}
\SNR\left(\prod_{i=1}^{p} s_i\right) \leq \SNR\left(\prod_{j=1}^{p-1} s_j\right), \forall p \in \{2, ...,n\}.
\end{equation}
\end{corollary}
\begin{proof}
The proof results immediately by induction from Theorem \ref{prop_snr}. Note that the inequality is strict when $\SNR(s_i)>0, \forall i \in \{1,2,...,p\}$.
\end{proof}

\begin{table*}[t]
\caption{Entropy after $k=6$ epochs for ResNet-18 on CIFAR-10, while alternating between the conventional and LeRaC training regimes.}
  \label{tab_entropy}
  \small{
  \begin{center}
  \begin{tabular}{|l|c|c|}
    \hline
    \multirow{2}{*}{Training Regime}     & \multicolumn{2}{c|}{Entropy} \\
    \cline{2-3}
    & First Conv Layer & Last Conv Layer \\
    \hline
    \hline
     conventional             & $0.9965$ & $0.9905$\\
      LeRaC (ours)              & $0.9970$ & $0.9968$ \\
    \hline
  \end{tabular}
    \end{center}
    }
      % \vspace{-0.3cm}
    
\end{table*}

\begin{figure*}[!t]
\begin{center}
\centerline{\includegraphics[width=0.95\linewidth]{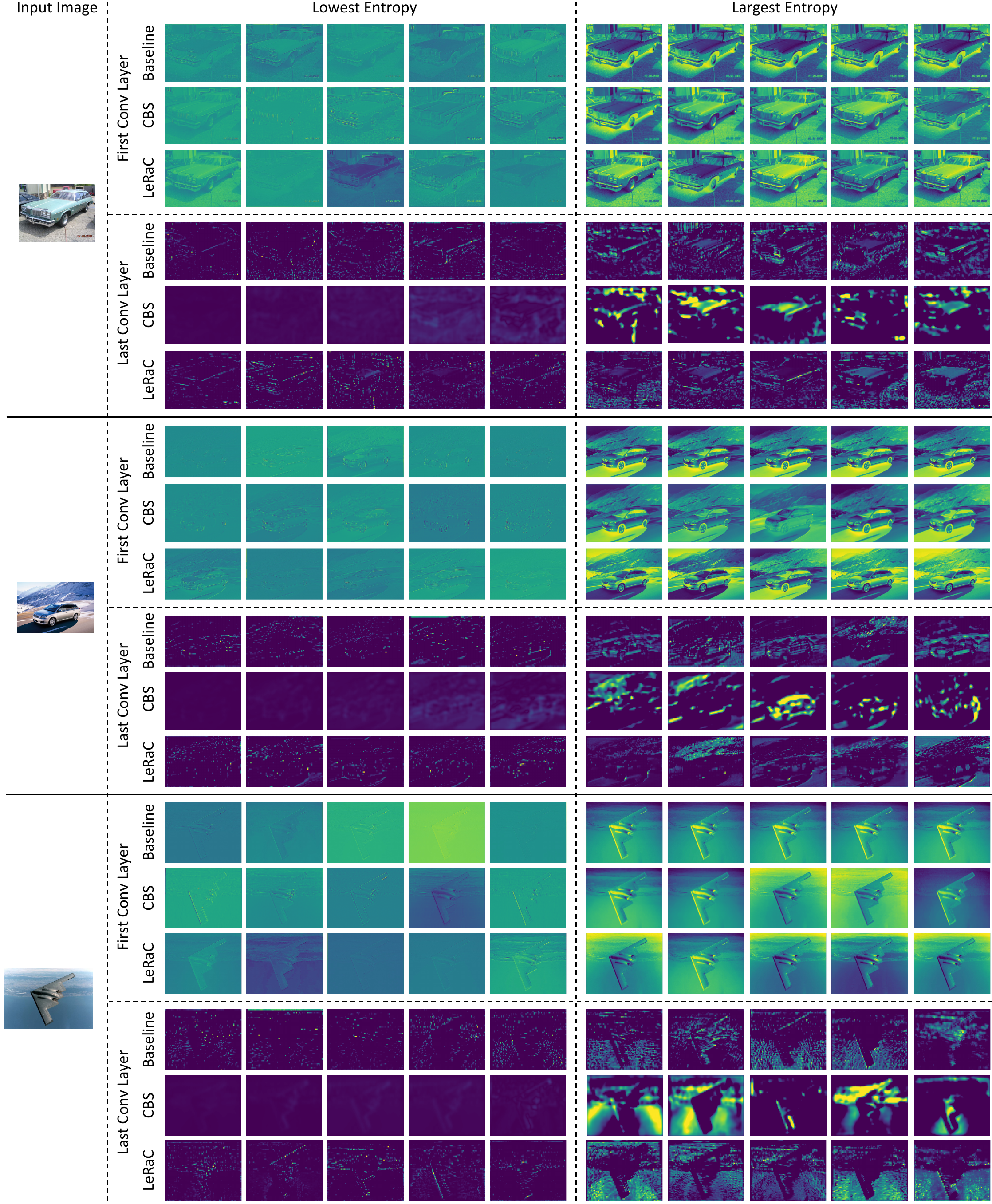}}
% \vspace{-0.25cm}
\caption{Activation maps with low and high entropy from the first and last conv layers of ResNet-18 trained on CIFAR-10 for $k=6$ epochs with the conventional (baseline) and LeRaC (ours) regimes. The input images are taken from ImageNet. Best viewed in color.}
\label{fig_activations}
%\vspace{-1.1cm}
\vspace{-0.2cm}
\end{center}
\end{figure*}

We employ Corollary \ref{prop_snr_multi} in the context of neural networks, where the input signal, which is expected to bear meaningful information and thus have a high SNR, is initially multiplied with random weights, which are expected to have low SNR values just after initialization. According to Corollary \ref{prop_snr_multi}, the SNR of the resulting signal (features) is gradually decreasing, layer by layer. In this context, we conjecture that optimizing the weights $\theta_i$ of layer $i$ to learn patterns from the signal (features) given as input to layer $i$ is suboptimal for layers that are sufficiently far away from the input. This happens because the respective features (passed to layer $i$) can contain a large amount of noise, which can derail the network towards adapting the weights $\theta_i$ to the noise instead of the clean signal. This phenomenon becomes more and more prevalent as the layer $i$ is placed farther away from the input. To regulate this phenomenon during the initial stages of the learning process, we propose to employ LeRaC and gradually decrease the learning rate as layers get deeper, allowing the network to optimize the earlier weights sooner. We underline that training the earlier layers also reduces the amount of noise in later layers, since the amount of noise in later layers is bounded by the amount of noise in earlier layers (according to Corollary \ref{prop_snr_multi}). As the amount of noise in later layers is progressively diminished, we can gradually increase the learning rates of later layers, allowing them to optimize their weights to cleaner signals (meaningful patterns).

\section{Empirical Proof}
\label{app_empirical_proof}

\noindent{\bf Noise quantification of early and later layers.}
The application of LeRaC is justified by the fact that the level of noise gradually grows with each layer during a forward pass through a neural network with randomly initialized weights. To empirically confirm this statement, we have computed the distances for the low-level (first conv) and high-level (last conv) layers between the activation maps at iteration $0$ (based on random weights) and the last iteration (based on weights optimized until convergence) for ResNet-18 on CIFAR-10, while using the conventional training regime. The computed distances shown in Table \ref{tab_noise_quant} confirm our conjecture, namely that shallow layers contain less noise than deep layers when applying the conventional training regime.

\begin{table*}[t]
\caption{Distances between feature maps at epoch $k=6$ and feature maps after the final epoch for ResNet-18 on CIFAR-10, while alternating between the conventional and LeRaC training regimes. Distances are independently computed for the first and last convolutional layers.}
  \label{tab_distance}
  \small{
  \begin{center}
  \begin{tabular}{|l|c|c|}
\hline
        \multirow{2}{*}{Training Regime}     & \multicolumn{2}{c|}{Distance} \\
        \cline{2-3}
    & First Conv Layer & Last Conv Layer \\
\hline
    \hline
         conventional             & $0.60$ & $0.37$\\
      LeRaC (ours)              & $0.61$ & $0.66$ \\
\hline
      \end{tabular}
    \end{center}
    }
      % \vspace{-0.3cm}
    
\end{table*}

\noindent{\bf Entropy of low-level versus high-level features.}
We show a few examples of training dynamics in Figure~\ref{fig_train_time}. All four graphs exhibit a higher gap between CBS and LeRaC in the first half of the training process, suggesting that LeRaC has an important role towards faster convergence. To assess the comparative quality of low-level versus high-level feature maps obtained either with conventional or LeRaC training, we compute the entropy of the first and last conv layers of ResNet-18 on CIFAR-10, after $k=6$ iterations. We report the computed entropy levels in Table~\ref{tab_entropy}. Conventional training seems to update deeper layers faster, observing a higher difference between the entropy levels of low-level and high-level features obtained with conventional training than with LeRac. This shows that LeRaC balances the training pace of low-level and high-level features. We conjecture that updating the deeper layers too soon could lead to overfitting to the noise still present in the early layers. This statement is supported by our empirical results on 12 data sets, showing that giving a chance to the early layers to converge before introducing large updates to the later layers leads to superior performance. 

Aside from computing the global entropy over all training samples, in Figure \ref{fig_activations}, we illustrate some activation maps with the highest and lowest entropy from the first and last conv layers for three randomly chosen examples from ImageNet. The activation maps are extracted at epoch $k=6$ from the ResNet-18 model trained on CIFAR-10 either with the conventional regime, the CBS regime or the LeRaC regime. In general, we observe that the low-level activation maps corresponding to LeRaC and CBS exhibit a higher degree of variability (being more distinct from each other), regardless of the entropy level (low or high). In the case of LeRaC, we believe the higher degree of variability comes from the fact that, having lower learning rates for the deeper layers, the model based on LeRaC is likely focused on finding a higher variety of patterns within the first layers to minimize the loss. Similarly, in the case of CBS, blurring the intermediary feature maps reduces the information propagated within the network. This compels the lower layers to identify and learn more distinctive patterns to minimize the loss. However, in general, the patterns found by LeRaC are more diverse. For instance, in the case of CBS, the low-level activation maps of the first image show greater similarity to each other, in contrast to those generated by LeRaC. For the third example (the image of an airplane), we observe that the activation maps with the highest entropy from the last conv layer produced by LeRaC have a higher entropy than the activation maps with the highest entropy produced by the conventional regime. This observation is in line with the results reported in Table~\ref{tab_entropy}, confirming that LeRaC is able to better balance the entropy of low-level and high-level features by preventing the faster convergence of the deeper layers. %, which induces a lower entropy of the high-level activation maps. 

\noindent{\bf Distances at epoch $\mathbf{k}$ versus final epoch.}
As discussed above, in Table~\ref{tab_entropy}, we report the entropy of the low-level and high-level layers after $k=6$ epochs, before and after using LeRaC to train ResNet-18 on CIFAR-10. However, we consider that using the distance to the final feature maps provides additional useful insights about how LeRaC works. To this end, we compute the Euclidean distances of both low-level and high-level features between epoch $k$ and the final epoch, before and after using LeRaC. We report the distances in Table~\ref{tab_distance}. The computed distances confirm our previous observations, namely that LeRaC is capable of balancing the training pace of low-level and high-level layers.

\section{Additional Experiments}
\label{app_additional_exp}
\vspace{-0.1cm}

\begin{table*}[t]
\caption{Average accuracy rates (in \%) over 5 runs for Wide-ResNet-50 on CIFAR-100 using different optimizers and training regimes (conventional versus LeRaC). The accuracy of the best training regime is highlighted in bold.}
  \label{tab_optimizer}
  \small{
  \setlength\tabcolsep{3.3pt}
  \begin{center}
  \begin{tabular}{|l|l|l|c|}
\hline
        Model  & Optimizer& Training Regime     & Accuracy \\
\hline
         % \multirow{3}{*}{ResNet-18} & Adam & conventional             & $57.90\!\pm\!0.21$ \\
         %                        & SGD & conventional             & $65.28\!\pm\!0.16$ \\
         %                        & SGD & LeRaC (ours)              & $\mathbf{66.02}\!\pm\!0.17$\\
\hline
        \multirow{3}{*}{Wide-ResNet-50} & Adam  & conventional             & $66.48\!\pm\!0.50$ \\
                                    & SGD   & conventional             & $68.14\!\pm\!0.16$ \\
                                    & SGD   & LeRaC (ours)              & $\mathbf{69.38}\!\pm\!0.26$\\
\hline
      \end{tabular}
    \end{center}
    }
      % \vspace{-0.3cm}
    
\end{table*}

% \begin{table}[t]
%   \small{
%   \renewcommand{\arraystretch}{1.1}
%   \setlength\tabcolsep{4.2pt}
%   \begin{center}
%   \begin{tabular}{|c|c|l|c|c|}
%     \hline    
%    \multirow{7}{*}{\rotatebox{90}{CREMA-D}} &   \multirow{7}{*}{\rotatebox{90}{SepTr}}   & conventional                   & $10^{-4}$-$10^{-4}$ &   $70.47\!\pm\!0.67$     \\
%     \cline{3-5}
%     &      & \multirow{6}{*}{LeRaC (ours)}                    & $10^{-2}$-$10^{-8}$ & $\mathbf{70.74}\!\pm\!0.55$ \\
%      &     &                     & $10^{-3}$-$10^{-8}$ & $\mathbf{70.91}\!\pm\!0.49$   \\
%     &     &                     & $10^{-4}$-$10^{-8}$ & $\mathbf{70.95}\!\pm\!0.56$   \\
%      &     &                     & $10^{-5}$-$10^{-8}$ & $70.32\!\pm\!0.57$   \\
%       &     &                     & $10^{-4}$-$10^{-7}$ & $\mathbf{70.49}\!\pm\!0.44$   \\
%        &     &                     & $10^{-4}$-$10^{-9}$ & $\mathbf{70.58}\!\pm\!0.48$   \\
%     \hline
%   \end{tabular}
%     \end{center}
%     }
%       \vspace{-0.3cm}
%     \caption{Average accuracy rates (in \%) over 5 runs for SepTr on CREMA-D, based on different ranges for the initial learning rates. The accuracy rates surpassing the baseline training regime are highlighted in bold.}
%   \label{tab_various_ranges}
% \end{table}

\begin{figure}[!t]
\begin{center}
\centerline{\includegraphics[width=1.0\linewidth]{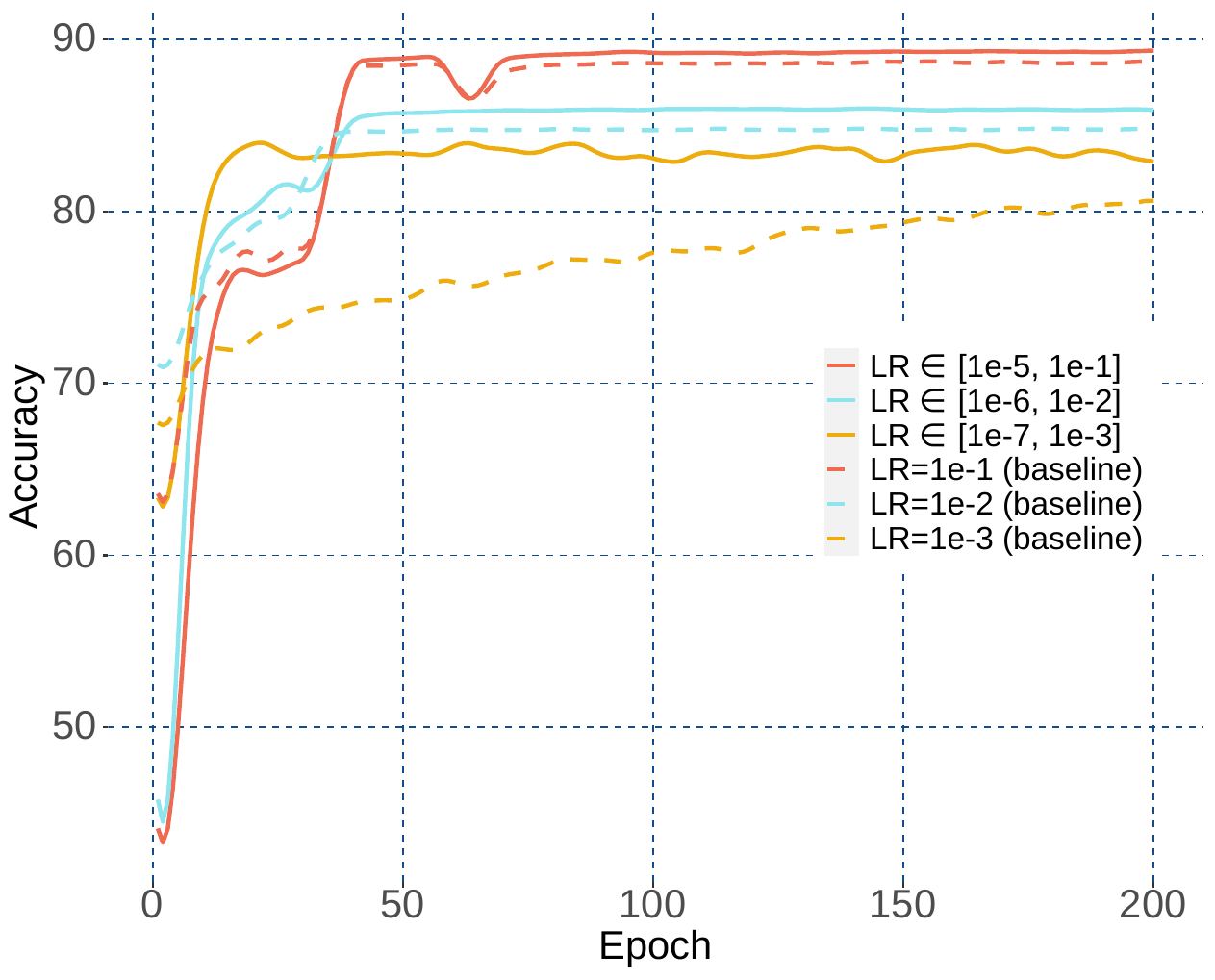}}
\caption{Test accuracy (on the y-axis) versus training time (on the x-axis) for ResNet-18 on CIFAR-10 with various configurations for the initial learning rates. Dashed lines correspond to the conventional regime, while continuous lines correspond to LeRaC. The different colors correspond to different initial learning rates. Best viewed in color.}
\label{fig_lr_var_r18_c10}
%\vspace{-1.1cm}
\end{center}
\end{figure}

\begin{figure}[!t]
\begin{center}
\centerline{\includegraphics[width=1.0\linewidth]{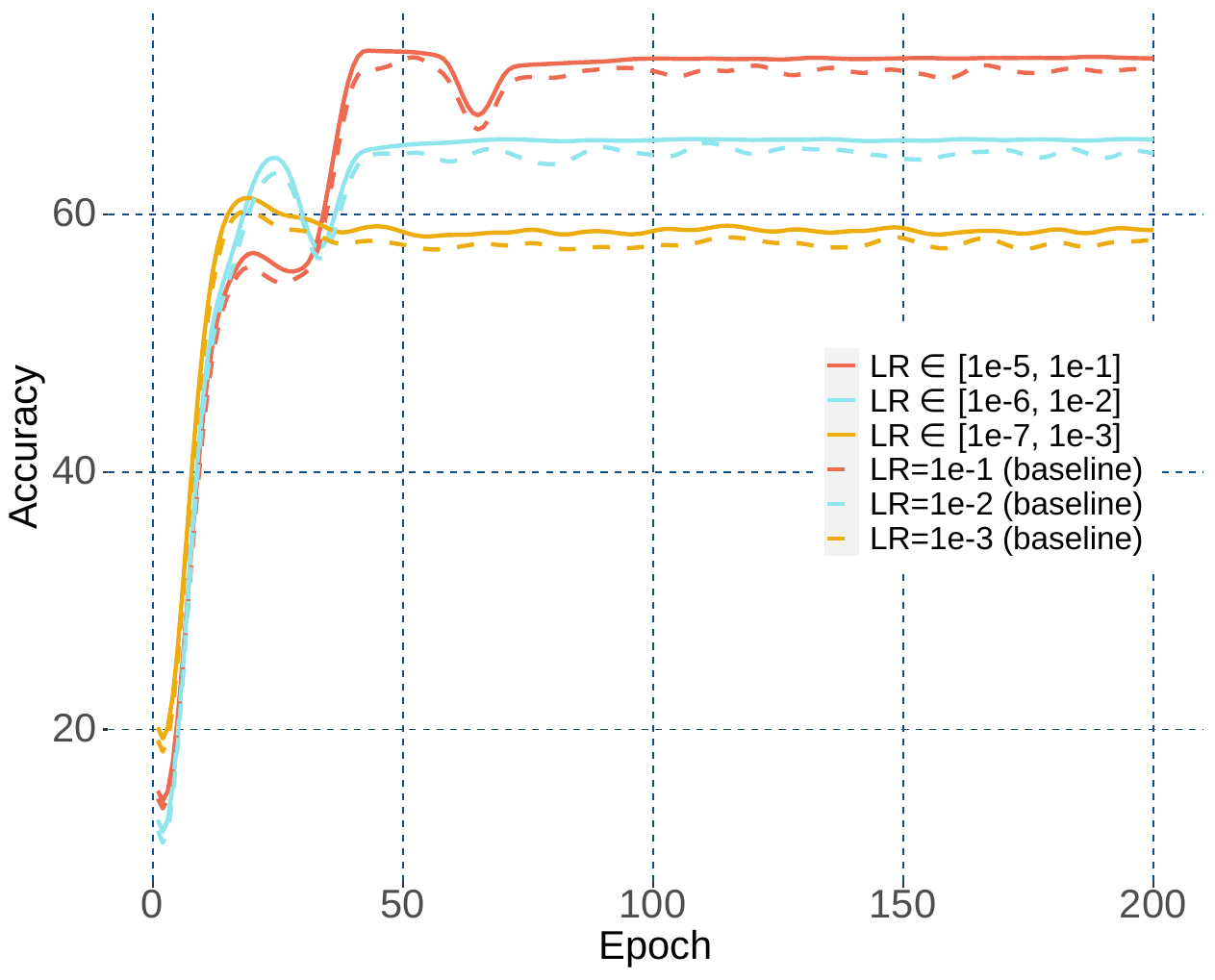}}
\caption{Test accuracy (on the y-axis) versus training time (on the x-axis) for ResNet-18 on CIFAR-100 with various configurations for the initial learning rates. Dashed lines correspond to the conventional regime, while continuous lines correspond to LeRaC. The different colors correspond to different initial learning rates. Best viewed in color.}
\label{fig_lr_var_r18_c100}
%\vspace{-1.1cm}
\end{center}
\end{figure}

\begin{figure}[!t]
\begin{center}
\centerline{\includegraphics[width=1.0\linewidth]{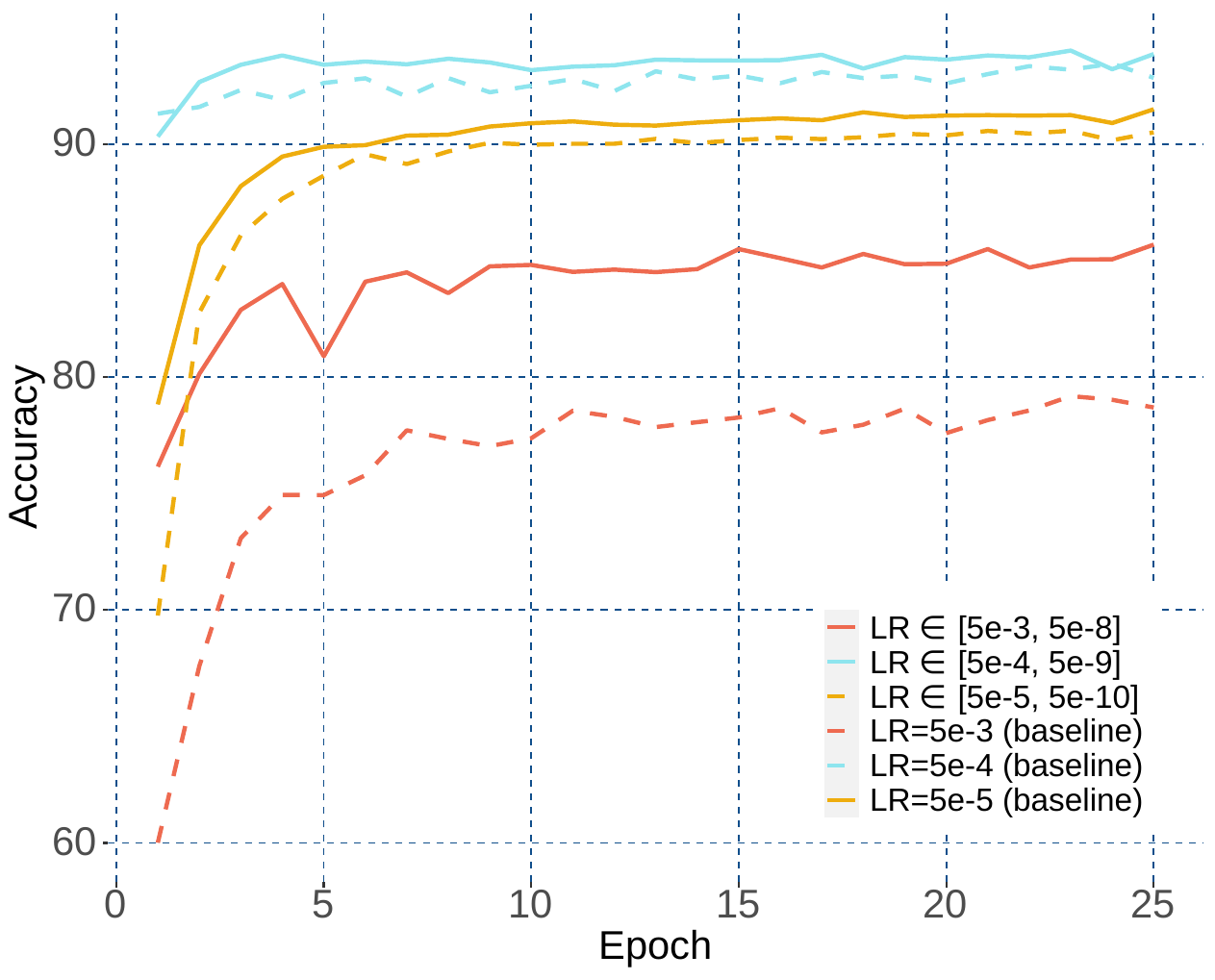}}
\caption{Test accuracy (on the y-axis) versus training time (on the x-axis) for the pre-trained CvT-13 on CIFAR-10 with various configurations for the initial learning rates. Dashed lines correspond to the conventional regime, while continuous lines correspond to LeRaC. The different colors correspond to different initial learning rates. Best viewed in color.}
\label{fig_lr_var_cvt_c10}
%\vspace{-1.1cm}
\end{center}
\end{figure}

\noindent{\bf Training progress for various initial learning rates.} We compare the training progress of the conventional and LeRaC training regimes. We first comparatively consider the progress of ResNet-18 on CIFAR-10, shown in Figure \ref{fig_lr_var_r18_c10}, and CIFAR-100, shown in Figure \ref{fig_lr_var_r18_c100}, respectively. LeRaC is consistently better than the conventional regime for all initial learning rate configurations, on both data sets. We next compare the progress on CIFAR-10 for ResNet-18, illustrated in Figure \ref{fig_lr_var_r18_c10}, and CvT-13 (pre-trained), illustrated in Figure \ref{fig_lr_var_cvt_c10}. The training progress of LeRaC is consistently above the training progress of the conventional regime, for both ResNet-18 and CvT-13. In summary, the results showcase the benefits on the training progress offered by LeRaC across distinct models and data sets.

% All our hyperparameters are either fixed without tuning or tuned on the validation data. In this ablation experiment, we present additional results with LeRaC using multiple ranges for $\eta_1^{(0)}$ and $\eta_n^{(0)}$ to demonstrate that LeRaC is sufficiently stable with respect to suboptimal hyperparameter choices. We carry out experiments with SepTr on CREMA-D. We report the corresponding results in Table~\ref{tab_various_ranges}. We observe that there are multiple hyperparameter configurations that lead to surpassing the baseline regime. This indicates that LeRaC can bring performance gains even outside the optimal learning rate bounds, demonstrating low sensitivity to suboptimal hyperparameter tuning.

\begin{table*}[!t]
 \caption{Average accuracy rates (in \%) over 5 runs on CIFAR-10, CIFAR-100 and Tiny ImageNet for CvT-13 based on different training regimes: conventional, CBS \citep{Sinha-NIPS-2020}, LeRaC with linear update, LeRaC with exponential update (proposed), and a combination of CBS and LeRaC.%\vspace{-0.45cm}
 }
  \label{tab_extra}
\small{
  \begin{center}
  \begin{tabular}{|l|l|c|c|c|}
    \hline
    Model       & Training Regime     & CIFAR-10  & CIFAR-100 & Tiny ImageNet \\
    \hline   
    \hline
    \multirow{4}{*}{CvT-13}             & conventional               &  $71.84 \pm 0.37$   &  $41.87 \pm 0.16$   & $33.38 \pm 0.27$ \\
    \cline{2-5}
                 & CBS %\citep{Sinha-NIPS-2020}
                 &  $72.64 \pm 0.29$   &  $44.48 \pm 0.40$   & $33.56 \pm 0.36$ \\
                    & LeRaC   &  $72.90 \pm 0.28$   &  $43.46 \pm 0.18$   & $33.95 \pm 0.28$ \\
    \cline{2-5}
                 & CBS %\citep{Sinha-NIPS-2020}
                 + LeRaC    &  $73.25 \pm 0.19$   &  $44.90 \pm 0.41$   & $34.20 \pm 0.61$ \\
    \hline
  \end{tabular}
  \end{center}
  }
  % \vspace{-0.3cm}
 
\end{table*}

\begin{table}[t]
\caption{Average accuracy rates (in \%) over 5 runs for ResNet-18 and Wide-ResNet-50 on CIFAR-100 using data augmentation and different training regimes (conventional versus LeRaC). The accuracy of the best training regime in each experiment is highlighted in bold.}
  \label{tab_vision_aug}
  \small{
  \setlength\tabcolsep{4.5pt}
  \begin{tabular}{|l|l|c|}
\hline
        Model  & Training Regime     & Accuracy \\
\hline
\hline
         \multirow{2}{*}{ResNet-18}              & conventional             & $72.25\!\pm\!0.04$ \\
                                &                LeRaC (ours)              & $\mathbf{73.51}\!\pm\!0.22$\\
\hline
        \multirow{2}{*}{Wide-ResNet-50}              & conventional             & $65.42\!\pm\!0.66$ \\
                                               & LeRaC (ours)              & $\mathbf{67.00}\!\pm\!0.55$\\
\hline
      \end{tabular}
    }
      % \vspace{-0.3cm}
    
\end{table}

\noindent{\bf SGD+LeRaC versus Adam.}
In Table \ref{tab_optimizer}, we present results showing that SGD and SGD+LeRaC obtain better accuracy rates than Adam \citep{Kingma-ICLR-1015} for the Wide-ResNet-50 model on CIFAR-100. This indicates that a simple optimizer combined with LeRaC can obtain better results than a state-of-the-art optimizer such as Adam. This justifies our decision to use a different optimizer for each neural model (see Table \ref{tab_parameters}).

\noindent{\bf Combining CBS and LeRaC.} 
Another interesting aspect worth studying is to determine if putting the CBS and LeRaC regimes together could bring further performance gains. We study the effect of combining CBS and LeRaC for CvT-13, since both CBS and LeRaC improve this model. In Table~\ref{tab_extra}, we present the results with CvT-13 on CIFAR-10, CIFAR-100 and Tiny ImageNet. The reported results show that the combination brings accuracy gains across all three data sets. We thus conclude that the combination of curriculum learning regimes is worth a try, whenever the two independent regimes boost performance.

\noindent{\bf Data augmentation on vision data sets.}
Following \citet{Sinha-NIPS-2020}, we did not use data augmentation for the vision data sets. We consider training data augmentation as an orthogonal method for improving results, expecting improvements for both baseline and LeRaC models. Nevertheless, since we extended the experimental settings considered in \citet{Sinha-NIPS-2020} to other domains, we took the liberty to use data augmentation in the audio domain (see the results in Table~\ref{tab_text}). The same augmentations (noise perturbation, time shifting, speed perturbation, mix-up and SpecAugment) are used for all audio models, ensuring a fair comparison. Moreover, we next present additional results with ResNet-18 and Wide-ResNet-50 on CIFAR-100 using the following augmentations: horizontal flip, rotation, solarization, blur, sharpening and auto-contrast. The results reported in Table \ref{tab_vision_aug} confirm that the performance gaps in the vision domain are in the same range after introducing data augmentation. In addition, we note that data augmentation seems to be rather harmful for the Wide-ResNet-50 model, which attains better results without data augmentation.

\begin{table}[t]
\caption{Average accuracy rates (in \%) over 5 runs for ResNet-18 on CIFAR-100 using limited training data (only 5\% of the full training set) and different training regimes: conventional, CBS \citep{Sinha-NIPS-2020} and LeRaC.  The accuracy of the best training regime is highlighted in bold.}
  \label{tab_limited_data}
  \small{
  \setlength\tabcolsep{3.3pt}
  \begin{tabular}{|c|l|c|}
\hline
        Training Set Size & Training Regime     & Accuracy \\
\hline
\hline
         \multirow{3}{*}{5\%}             & conventional             & $23.86 \pm 0.32$ \\
                                                     & CBS %\citep{Sinha-NIPS-2020}
                                                     &  ${24.79} \pm 0.17$ \\
                                &                LeRaC (ours)              & $\mathbf{25.04} \pm 0.22$\\
\hline
      \end{tabular}
    }
      % \vspace{-0.3cm}
    
\end{table}

\noindent{\bf Limited data regime.}
In all our experiments carried out so far, the evaluated models were trained on the complete training sets. However, it is interesting to find out how our strategy behaves in a limited data regime. To this end, we conduct another experiment to compare LeRaC with the conventional and CBS regimes in a limited data scenario, considering only 5\% of the training data. We present the results for ResNet-18 on CIFAR-100 in Table \ref{tab_limited_data}. The  results indicate that LeRaC keeps its performance edge in the limited data regime. We therefore conclude that LeRaC can also be useful when limited training data is available.

%\end{appendices}

{\small
\bibliography{references}
\bibliographystyle{sn-vancouver}
}

%%%%%%%%%%%%%%%%%%%%%%%%%%%%%%%%%%%%%%%%%%%%%%%%%%%%%%%%%%%%

\end{document}

%% file: math_commands.tex
\usepackage{amsmath,amsfonts,bm}

\def\eqref#1{equation~\ref{#1}}
\def\1{\bm{1}}

\DeclareMathAlphabet{\mathsfit}{\encodingdefault}{\sfdefault}{m}{sl}
\SetMathAlphabet{\mathsfit}{bold}{\encodingdefault}{\sfdefault}{bx}{n}

\newcommand{\Var}{\mathrm{Var}}